\documentclass{article}       % onecolumn (second format)
\usepackage[a4paper, total={6in, 8in}]{geometry}

\usepackage{enumerate}
\usepackage[misc]{ifsym}
\usepackage{lineno}
\usepackage{textcomp}
\usepackage{graphicx}
\usepackage{amsmath,amssymb,amsfonts}
\usepackage{pgfplots}
\pgfplotsset{compat=1.15}
\usetikzlibrary{arrows}
\usepackage{algorithm} 
\usepackage{algpseudocode}
\usepackage[caption=false,font=normalsize,labelfont=sf,textfont=sf]{subfig}
\usepackage{lipsum} 
\usepackage{url}
\usepackage{authblk}

\usepackage{lineno}
\usepackage[semicolon,authoryear]{natbib}
\begin{document}

\sloppy

\title{Robust optimal well control using an adaptive multi-grid reinforcement learning framework}

\author[1]{Atish Dixit}
\author[2]{Ahmed H. ElSheikh}
\affil[1,2]{School of Energy, Geoscience, Infrastructure and Society \protect\\
Heriot-Watt University}
\affil[1]{\textup{Email: ad181@hw.ac.uk}}
\affil[2]{\textup{Email: a.elsheikh@hw.ac.uk}}
\date{}                     %% if you don't need date to appear
\setcounter{Maxaffil}{0}
\renewcommand\Affilfont{\itshape\small}

\maketitle

\begin{abstract}
Reinforcement learning (RL) is a promising tool to solve robust optimal well control problems where the model parameters are highly uncertain, and the system is partially observable in practice. However, RL of robust control policies often relies on performing a large number of simulations. This could easily become computationally intractable for cases with computationally intensive simulations. To address this bottleneck, an adaptive multi-grid RL framework is introduced which is inspired by principles of geometric multi-grid methods used in iterative numerical algorithms. RL control policies are initially learned using computationally efficient low fidelity simulations using coarse grid discretization of the underlying partial differential equations (PDEs). Subsequently, the simulation fidelity is increased in an adaptive manner towards the highest fidelity simulation that correspond to finest discretization of the model domain. The proposed framework is demonstrated using a state-of-the-art, model-free policy-based RL algorithm, namely the Proximal Policy Optimisation (PPO) algorithm. Results are shown for two case studies of robust optimal well control problems which are inspired from SPE-10 model 2 benchmark case studies. Prominent gains in the computational efficiency is observed using the proposed framework saving around 60-70\% of computational cost of its single fine-grid counterpart.
% \keywords{reinforcement learning \and adaptive \and multi-grid framework \and robust \and optimal control}
\end{abstract}

%%%%%%%%%%%%%%%%%%%%%%%%%%%%%%%%%%%%%
%%%%%%%%%%%%%%%%%%%%%%%%%%%%%%%%%%%%%
\section{Introduction}
Optimal control problem involves finding controls for a dynamical system such that a certain objective function is optimized over a pre-defined simulation time. 
Recently, reinforcement learning (RL) has been demonstrated as an effective method to solve stochastic optimal control problems in fields like manufacturing \citep{dornheim2020model}, energy \citep{anderlini2016control} and fluid dynamics \citep{rabault2019artificial}. 
RL, being virtually a stochastic optimisation method, involves a huge number of exploration and exploitation attempts in order to learn the optimal control policy. 
As a result, learning the optimal policy requires a large number of simulations of the controlled dynamical system which is often computationally expensive.
In this paper, an adaptive multi-grid RL framework is introduced to reduce overall computational cost of number of simulations required to learn the optimal control policy. 
% This framework is essentially inspired by the principles of geometric multi-grid methods used in iterative numerical algorithms. 
% The optimal policy learning process is initiated using a low fidelity simulation that correspond to a coarse grid discretization of the underlying partial differential equations (PDEs). 
% This learned policy is then resused to further train it using high fidelity simulations in an adaptive and incremental manner.
% Robustness of the policy learned using this framework is finally evaluated against uncertainties in the model dynamics.

Various research studies have shown the effectiveness of using multi-grid method to improve the convergence rate of reinforcement learning. 
\cite{anderson1994multigrid} extend Q-Learning by casting it as a multi-grid method and has shown a reduction in the number of updates required to reach a given error level in the Q-function.
\cite{ziv2005multigrid} and \cite{pareigis1996multi} formulated the value function learning process with a Hamilton-Jacobi-Bellman (HJB) equation which is solved using algebraic multi-grid methods.
Albeit the effectiveness of this strategy, HJB formulation is only feasible when the model dynamics are well defined. As a result, these methods cannot to be applied to problems where the model dynamics are an approximate representation of reality.
\cite{li2015multi} used multi-grid approach to compute tabular Q values for energy conservation and comfort of HVAC in buildings which is applicable to certain simple RL problems with finite and discrete state-action space.
In this paper, the aim is to present a generalized multi-grid RL approach which can be applied on both, discrete and continuous, state and action space where HJB formulation may not be possible for instance, when the transition in model dynamics is not necessarily differentiable and/or when the model is stochastic.
This framework is essentially inspired by the principles of geometric multi-grid methods used in iterative numerical algorithms. 
The optimal policy learning process is initiated using a low fidelity simulation that correspond to a coarse grid discretization of the underlying partial differential equations (PDEs). 
This learned policy is then reused to further train it using high fidelity simulations in an adaptive and incremental manner.
Robustness of the policy learned using this framework is finally evaluated against uncertainties in the model dynamics.

In reinforcement learning literature, such a learning process is categorized as transfer learning. 
The idea behind transfer learning is that instead of learning directly on the target task, the agent can first train on one or more source task(s), and transfer the knowledge acquired to aid in solving the target task \citep{taylor2009transfer}.
In the context of current study, highest fidelity simulation correspond to the target task which is assumed to have the fine-grid discretization which guarantees good approximation of the output quantities of interest with the accuracy required by the problem at hand.
Low grid fidelity simulations that compromises on the accuracy of these quantities, on the other hand, correspond to source tasks.
These low grid fidelity simulations are generated using a degree of freedom parameter called grid fidelity factor (much like in the study done by \cite{narvekar2016source}).
Transfer learning is a much broader sub-domain of RL that covers knowledge transfer in the form of data samples \citep{lazaric2008transfer}, policies \citep{fernandez2010probabilistic}, models \citep{fachantidis2013transferring} or value functions \citep{taylor2005behavior}.
In this study, the knowledge transfer is done in the form of the policy for a model-free, on-policy algorithm called proximal policy optimisation (PPO). 
Since the policy is designed for the state and actions corresponding to the highest fidelity simulation we employ a mapping function that maps states and actions from low fidelity simulations to high fidelity simulations and vice versa. 
This is done by defining restriction (mapping from high to low fidelity simulation) and prolongation (mapping from low to high fidelity simulation) operators which are normally found in classical geometric multi-grid methods.

Effectiveness of this multi-grid RL framework is demonstrated for robust optimal well control problem which is a subject of intensive research activities in subsurface reservoir management \citep{van2009robust, roseta2004robust, brouwer2001recovery}.
For this problem, the dynamical system under consideration is non-linear and, in practice, is partially observable since the data is only available at a sparse set of points (i.e. well locations). 
Furthermore, the subsurface model parameters are highly uncertain due to sparsity of the available field data.
Optimal well control problem consists of optimizing the control variables like valve openings of wells in order to maximize sweep efficiency of injector fluid throughout the reservoir life. 
Reservoir permeability field is considered as an uncertain model parameter for which the uncertainty distribution is known. 
Although the proposed framework is demonstrated for robust optimal well control problem, it is designed to be general enough to be applicable to similar optimal control problems governed by a set of PDEs.
Two test cases -- both representing a distinct model parameter uncertainty and control dynamics -- are used to demonstrate the computational gains of using the multi-grid idea.

The outline of the rest of this paper is as following: Section \ref{sect: methodology} provides the problem description and proposed framework to solve robust optimal well control problem.
Section \ref{sect: case studies} details the model parameters for the two case studies designed for demonstration. 
Results of the proposed framework on these two case studies are demonstrated in section \ref{sect: results}. Finally, section \ref{sect: conclusion} concludes with the research study summary and an outlook of future research directions.

\section{Methodology}
\label{sect: methodology}
%\subsection{problem description}
Fluid flow control in subsurface reservoirs has many engineering applications ranging from the financial aspects of efficient hydrocarbon production to the environmental problems of contaminated removal from polluted aquifers \citep{whitaker1999single}.
In this paper, a canonical single-phase subsurface flow control problem (also referred as robust optimal well control problem) is studied where water is injected in porous media to displace a contaminant. 
This process is commonly modeled using an advection equation for tracer flow through porous media (also referred as Darcy flow through porous media) over the temporal domain $\mathcal{T} = [t_0, t_M] \subset \mathbb{R}$ and spatial domain $\mathcal{X} \subset \mathbb{R}^2$.
In the context of fluid displacement (e.g. groundwater decontamination), the tracer corresponds to clean water injected in the reservoir from the injector wells and the non-traced fluid corresponds to the displaced contaminated water from the reservoir through producer wells. 
The source and sink locations within the modeled domain correspond to injector and producer wells, respectively.
The tracer flow models water flooding with the fractional variable $s(x,t) \in [0,1]$ (also referred as saturation) which represents the fraction of injected clean water to the displaced contaminated water at location $x \in \mathcal{X}$ and time $t \in \mathcal{T}$. 
The fluid flow in and out of the domain is represented with $a(x,t)$ which is treated as source/sink terms of the governing equation.
Set of well locations are denoted as $x' \in \mathcal{X'}$ (where $\mathcal{X'} \subset \mathcal{X}$).
In other words, $a(x,t)$ is assigned to zero everywhere in the domain $\mathcal{X}$ except the set of locations $x'$.
The controls $a^+(x,t)$ (formulated as $\max(0,a(x,t))$) and $a^-(x,t)$ (formulated as $\min(0,a(x,t))$) represent the injector and producer flow controls, respectively (note that $a=a^+ + a^-$).
Task of the problem under consideration, 
is to find optimal controls $a^*(x',t)$ which is the solution of following closed-loop optimisation problem:
\begin{subequations}
\begin{align}
    & \max_{s(\cdot), a(\cdot)} \int_{t_0}^{t_M}  \left ( \sum_{x'}a^-(x',t)(1-s(x',t))  \right )  dt, & x' \in \mathcal{X'}, \ t \in \mathcal{T} \label{eq: obj_fun} \\
    & \frac{ds}{dt} = \frac{1}{\phi} \left (a^+ + sa^- - \nabla \cdot sv \right ), & x \in \mathcal{X}, \ t \in \mathcal{T} \label{eq: gov_eq} \\
    & s(\cdot,t_0) = s_0,\ \ v \cdot \textbf{n} = 0, & \label{eq: init_eq}\\
    & \sum_{x'} a^+(x',t)  = -\sum_{x'} a^-(x',t) = c, & x' \in \mathcal{X'}, \ t \in \mathcal{T} \label{eq: constr}
\end{align}
\label{eq: prob_def}
\end{subequations}

The objective function defined in equation \eqref{eq: obj_fun} represents the total displaced fluid flow out of the reservoir (e.g. contaminated water production) and is maximized on the finite time interval $\mathcal{T}$. 
The intigrand in this function is referred as Lagrangian term in control theory and is often denoted by $L(s,a)$. 
The water flow trajectory $s(x,t)$, is governed by advection equation \eqref{eq: gov_eq} which is solved given the velocity field $v$, which is obtained from the Darcy's law: $v = - (k/\mu) \nabla p$. The pressure $p(x,t) \in \mathbb{R}$, is obtained from the pressure equation, $-\nabla \cdot (k/\mu) \nabla p = a $. 
Porosity $\phi (x,\cdot)$, permeability $k (x,\cdot)$, and viscosity $\mu (x,\cdot)$, are the model parameters.
Permeability $k$, represents the model uncertainty and is treated as a random variable that follows a known probability density function $\mathcal{K}$ with $K$ as its domain. 
The initial and no flow boundary conditions are defined in equation \eqref{eq: init_eq}, where \textbf{n} denotes outward normal vector from the boundary of $\mathcal{X}$. 
The constraint defined in equation \eqref{eq: constr} represent the fluid incompressibility assumption along with the fixed total source/sink term $c$ which represents total water injection rate in the reservoir.
In a nutshell, the optimisation problem provided in equations \eqref{eq: prob_def} is solved to find the optimal controls $a^*(x',t)$ such that they are robustly optimal over the entire permeability uncertainty domain, $K$.
% The controls $a(x',t)$ are discretized on time interval $t_0<t_1<\cdots <t_{m}<t_{m+1}=T$.

\subsection{RL framework}
According to RL convention, the optimal control problem defined in equation \eqref{eq: prob_def} is modeled as a Markov decision process which is defined as a quadruple $\left \langle \mathcal{S},\mathcal{A},\mathcal{P},\mathcal{R} \right \rangle$. Here, $\mathcal{S} \subset \mathbb{R}^{n_s}$ is set of all possible states with the dimension $n_s$, $\mathcal{A} \subset \mathbb{R}^{n_a}$ is a set of all possible actions with the dimension $n_a$. 
The state $S$, is represented with the saturation $s(x,\cdot)$ and pressure $p(x,\cdot)$ values over the entire domain $\mathcal{X}$. 
The action $A$, is represented with an array of well control values $a(x',\cdot)$. 
More details of this array like representation of action are presented in section \ref{sect: rl_formulation}.
The optimal control problem defined in equation \eqref{eq: prob_def} is discretized into $M$ control steps and as a result, its solution is a set of optimal control values ${a^*(x',t_1), a^*(x',t_2), \ldots, a^*(x',t_M)}$ where $t_0<t_1<t_2<\cdots <t_{M}$.
The transition function
$\mathcal{P}:\mathcal{S}\times \mathcal{A} \rightarrow \mathcal{S}$, is assumed to follow Markov property.
That is, transition to the state $S(t_{m+1})$ is obtained by executing the actions $A(t_{m})$ when in the state $S(t_{m})$.
Such transition function is obtained by discretizing equation \eqref{eq: gov_eq}.
For a transition from the state $S(t_{m})$, to the state $S(t_{m+1})$, the real valued reward $R(t_{m+1})$, is calculated as $R(t_{m+1})=\mathcal{R}(S(t_{m}), A(t_{m}), S(t_{m+1}))$, where $\mathcal{R}:\mathcal{S} \times \mathcal{A} \times \mathcal{S} \rightarrow \mathbb{R}$ is the reward function.
The reward function is obtained by discretizing the objective function (equation \eqref{eq: obj_fun}) into control steps such that, 
\begin{equation}
    R(t_{m+1})= \int_{t_m}^{t_{m+1}}  L(s,a) dt.
    \label{eq: reward_func}
\end{equation}
The optimal controls are obtained by learning a control policy function which is defined as $\pi:\mathcal{S} \rightarrow \mathcal{A}$. This function is denoted as $\pi(A|S)$ and is generally represented with a neural network. Essentially, the control policy $\pi(A|S)$, maps a given state $S(t_m)$, into an action $A(t_m)$.
For an optimal control problem, with $M$ control steps,
the goal of reinforcement learning is to find an optimal policy $\pi^*(A|S)$ such that the expected reward $G = \sum_{m=1}^M \gamma^{m-1} R(t_{m})$, is maximized. 
Note that immediate rewards $R$, are exponentially decayed by the discount rate $\gamma \in [0,1]$. 
The discount rate represents how myopic the learned policy is, for instance, learned policy is considered completely myopic when $\gamma=0$.  
The controller, which is also referred to as an agent, follows the policy and explores various control trajectories by interacting with the environment which consists of a transition function $\mathcal{P}$ and a reward function $\mathcal{R}$.
The data gathered by these control trajectories are used to update the policy towards optimality.
Each such update of the policy is referred to as the policy iteration.
In RL literature, a single complete control trajectory is referred to as an episode.
Essentially, RL algorithms attempt to learn the optimal policy $\pi^*(A|S)$ from a randomly initialized policy $\pi(A|S)$, by exploring state-action space by executing a high number of episodes. 

In order to represent the variability in permeability, a finite number $l$, of well spread uncertainty distribution samples is chosen. 
This is achieved with a clustering analysis (please refer appendix \ref{app: cluster} for cluster analysis formulation used in this paper) of the domain $K$. 
The sample vector $\textbf{k}= \{k_1, k_2, \cdots k_l\}$, is constructed with samples of the distribution $\mathcal{K}$, which are located nearest to the cluster centers.
The policy $\pi^*(A|S)$, is learned by randomly selecting the parameter $k$ from the training vector $\textbf{k}$ at the beginning of every episode. 
The policy return $R^{\pi(A|S)}$, is computed by averaging the returns of policy $\pi(A|S; k_i)$ (policy applied on the simulation where permeability is set to $k_i$) on $l$ simulations, which is formulated as, 
\begin{equation}
    R^{\pi(A|S)} = \frac{1}{l}  \sum_{i=1}^l \sum_{m=0}^{M-1} \int_{t_m}^{t_{m+1}} L(s, \pi(A|S; k_i)) dt.
\end{equation}
In optimal well control problems, the system is partially observable, that is, reservoir information is only available at well locations throughout the reservoir life cycle. 
In order to accommodate this fact, the agent is provided with the available observation as its state. For this study, observation is represented with a set of saturation and pressure values at the well locations $x'$. 
Note that, with such representation of states, the underlying assumption of Markov property of the transition function is approximated. 

\subsection{Learning convergence criteria}
\begin{figure}
    \centering
    \begin{tabular}{c}
        \subfloat[effect of $\delta$ on convergence criteria]{\includegraphics[width=0.9 \columnwidth]{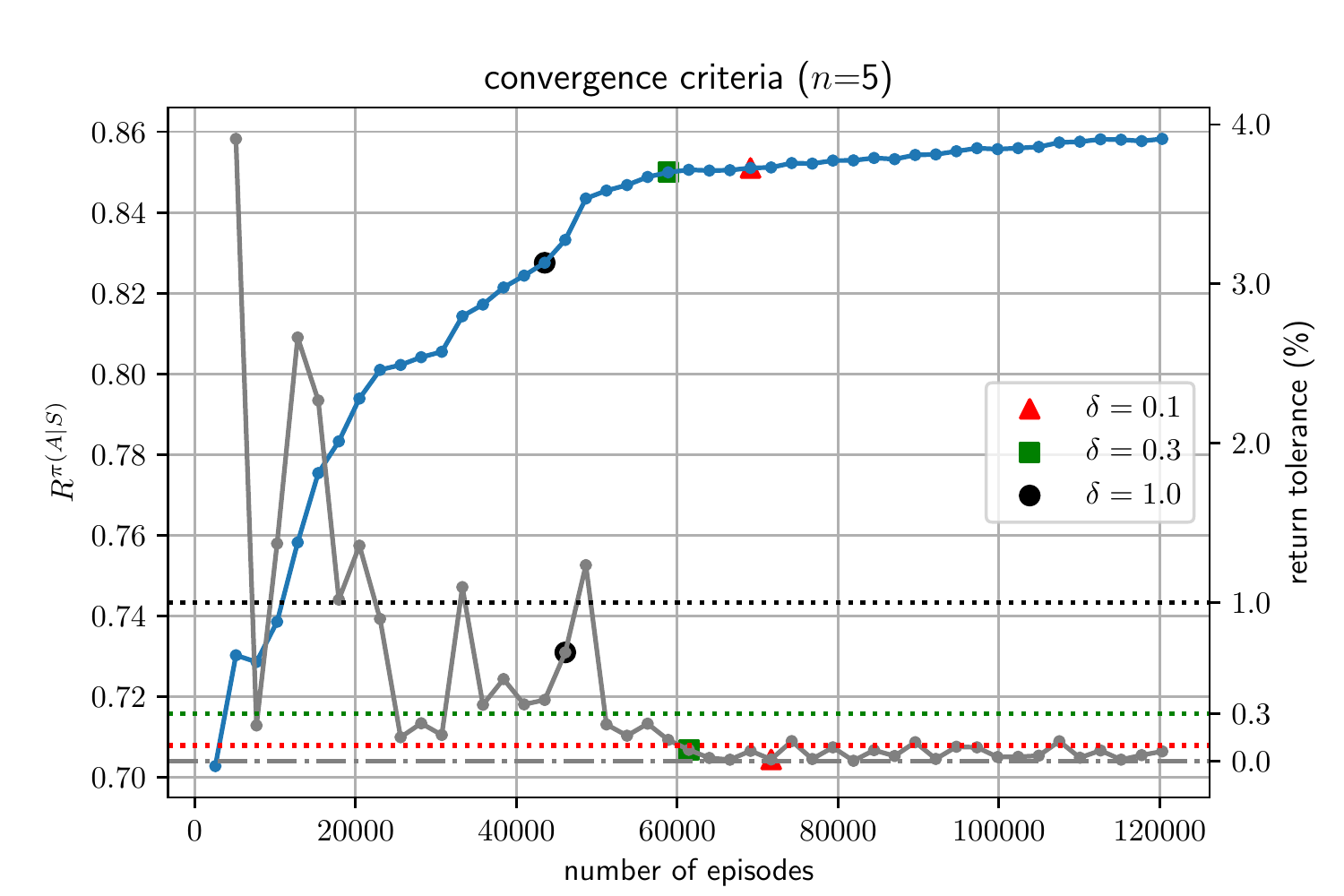}} \\
        \subfloat[effect of $n$ on convergence criteria]{\includegraphics[width=0.9 \columnwidth]{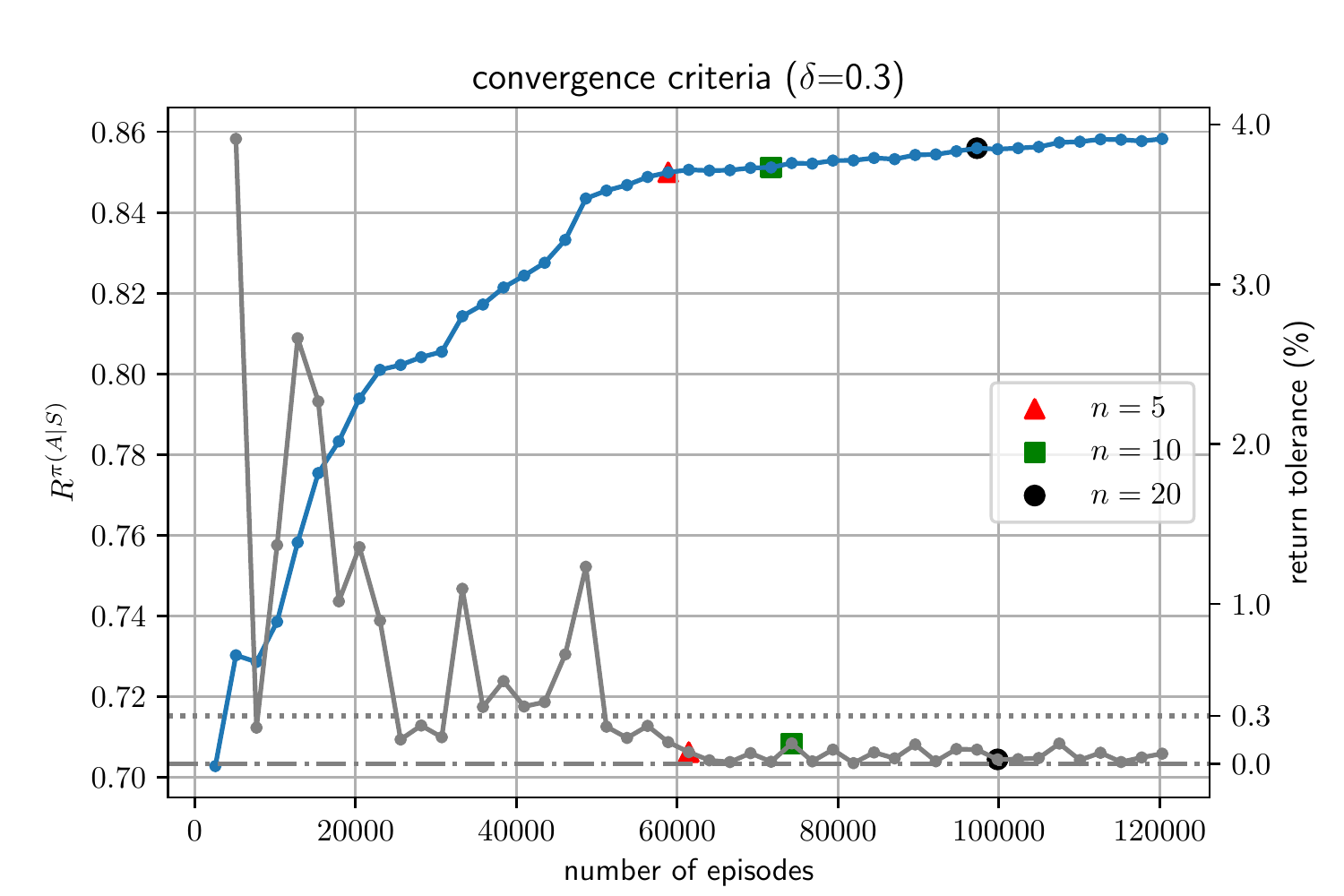}}
    \end{tabular}
    \caption{Plot of policy returns versus number of training episodes}
    \label{fig:conv_criteria}
\end{figure}

The optimal policy convergence is detected by monitoring the policy return $R^{\pi(A|S)}$, after every policy iteration. 
Conventionally, when this value converges to a maximum value, the optimal policy is assumed to be learned. 
The convergence criteria for $i$th policy iteration is defined as,
\begin{equation}
    \delta_i =  \left | \frac{R^{\pi(A|S)}_i - R^{\pi(A|S)}_{i-1} }{\max (R^{\pi(A|S)}_{i-1}, \epsilon) }  \right | < \delta,
    \label{eq: conv_cri}
\end{equation}
where $\delta_i$ is the return tolerance at $i$th policy iteration, $\delta$ is stopping tolerance and $\epsilon$ is a small non-zero number used to avoid division by zero.
The convergence of policy learning is often flat near the optimal result.
For this reason, the convergence criteria defined in equation \eqref{eq: conv_cri} is checked for the latest $n$ consecutive policy iterations.
For instance, if $\textbf{r}$ is the array of monitored values of $R^{\pi(A|S)}$ at all policy iterations, the policy $\pi(A|S)$ is considered converged when the convergence criteria (equation \eqref{eq: conv_cri}) for last $n$ policy iterations is met. 
Algorithm \ref{alg:isconverged} delineates the pseudocode for this convergence criteria.
\begin{algorithm*}
	\caption{learning convergence criteria} 
	\begin{algorithmic}[1]
	\Procedure{IsConverged}{$\textbf{r}$, $n$, $\delta$}
	\If {$length(\textbf{r}) < n$}
	    \Return {False}
	\EndIf
    \State compute $\delta_i$ (equation \eqref{eq: conv_cri}) for last $n$ values of \textbf{r} and get its maximum $\delta_{max}$
    \If{$\delta_{max} < \delta$}
        \State \Return{True}
    \Else 
        \State \Return{False}
    \EndIf
    \EndProcedure
	\end{algorithmic} 
	\label{alg:isconverged}
\end{algorithm*}
Figure \ref{fig:conv_criteria} illustrates effect of $n$ and $\delta$ on convergence criteria for an example of reinforcement learning process. Policy return plot is shown in blue color where each value at policy iteration is shown with a dot. The corresponding return tolerance is plotted in gray color which is represented in percentage format ($\delta_i\times100$, where $\delta_i$ is computed from equation \eqref{eq: conv_cri}). It can be seen that the convergence criteria (denoted with markers on these plots) is more stringent when the stopping tolerance $\delta$, is smaller and consecutive policy iteration steps $n$, are higher.

\subsection{Adaptive multi-grid RL framework}
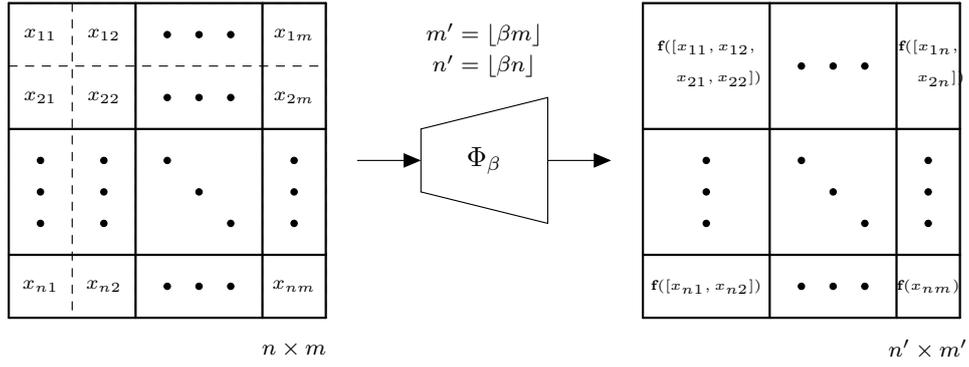
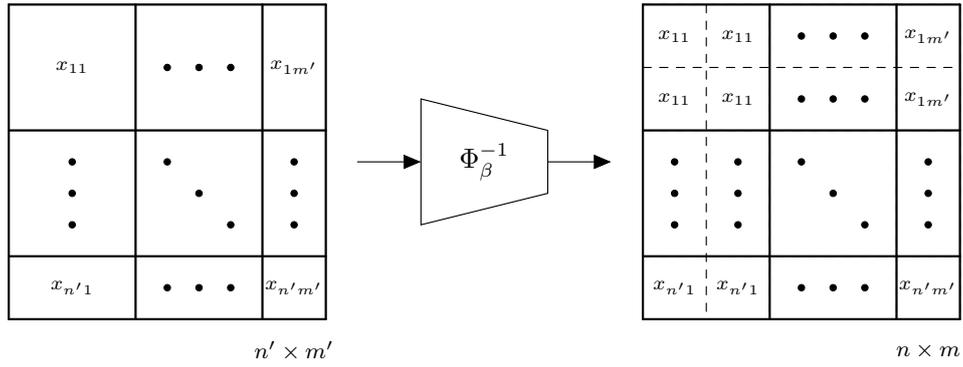
\begin{figure}
    \centering
    \begin{tabular}{c}
        \subfloat[restriction operator, $\Phi_{\beta}$]{\resizebox{0.9\columnwidth}{!} {
        \begin{tikzpicture}[line cap=round,line join=round,>=triangle 45,x=1cm,y=1cm, scale=0.8]
\clip(-8,-4) rectangle (8,3);

% fine grid left
\draw [dashed, line width=0.4pt] (-7.5,2.5)-- (-7.5,-2.5);
\draw [dashed, line width=0.4pt] (-6.5,2.5)-- (-6.5,-2.5);
\draw [dashed, line width=0.4pt] (-5.5,2.5)-- (-5.5,-2.5);
\draw [dashed, line width=0.4pt] (-3.5,2.5)-- (-3.5,-2.5);
\draw [dashed, line width=0.4pt] (-2.5,2.5)-- (-2.5,-2.5);

\draw [dashed, line width=0.4pt] (-7.5,2.5)-- (-2.5,2.5);
\draw [dashed, line width=0.4pt] (-7.5,1.5)-- (-2.5,1.5);
\draw [dashed, line width=0.4pt] (-7.5,0.5)-- (-2.5,0.5);
\draw [dashed, line width=0.4pt] (-7.5,-1.5)-- (-2.5,-1.5);
\draw [dashed, line width=0.4pt] (-7.5,-2.5)-- (-2.5,-2.5);

% coarse grid left
\draw [line width=0.8pt] (-7.5,2.5)-- (-7.5,-2.5);
\draw [line width=0.8pt] (-5.5,2.5)-- (-5.5,-2.5);
\draw [line width=0.8pt] (-3.5,2.5)-- (-3.5,-2.5);
\draw [line width=0.8pt] (-2.5,2.5)-- (-2.5,-2.5);

\draw [line width=0.8pt] (-7.5,2.5)-- (-2.5,2.5);
\draw [line width=0.8pt] (-7.5,0.5)-- (-2.5,0.5);
\draw [line width=0.8pt] (-7.5,-1.5)-- (-2.5,-1.5);
\draw [line width=0.8pt] (-7.5,-2.5)-- (-2.5,-2.5);

% labels left
\draw (-7,2) node[anchor=center] {\scriptsize{$x_{11}$}};
\draw (-6,2) node[anchor=center] {\scriptsize{$x_{12}$}};
\draw (-3,2) node[anchor=center] {\scriptsize{$x_{1m}$}};

\draw (-7,1) node[anchor=center] {\scriptsize{$x_{21}$}};
\draw (-6,1) node[anchor=center] {\scriptsize{$x_{22}$}};
\draw (-3,1) node[anchor=center] {\scriptsize{$x_{2m}$}};

\draw (-7,-2) node[anchor=center] {\scriptsize{$x_{n1}$}};
\draw (-6,-2) node[anchor=center] {\scriptsize{$x_{n2}$}};
\draw (-3,-2) node[anchor=center] {\scriptsize{$x_{nm}$}};

\draw (-3,-3) node[anchor=center] {\footnotesize{$n\times m$}};

% dots left
\draw [fill=black] (-4,2) circle (0.05cm);
\draw [fill=black] (-5,2) circle (0.05cm);
\draw [fill=black] (-4.5,2) circle (0.05cm);

\draw [fill=black] (-4,1) circle (0.05cm);
\draw [fill=black] (-5,1) circle (0.05cm);
\draw [fill=black] (-4.5,1) circle (0.05cm);

\draw [fill=black] (-4,-2) circle (0.05cm);
\draw [fill=black] (-5,-2) circle (0.05cm);
\draw [fill=black] (-4.5,-2) circle (0.05cm);

\draw [fill=black] (-7,0) circle (0.05cm);
\draw [fill=black] (-7,-1) circle (0.05cm);
\draw [fill=black] (-7,-0.5) circle (0.05cm);

\draw [fill=black] (-6,0) circle (0.05cm);
\draw [fill=black] (-6,-1) circle (0.05cm);
\draw [fill=black] (-6,-0.5) circle (0.05cm);

\draw [fill=black] (-3,0) circle (0.05cm);
\draw [fill=black] (-3,-1) circle (0.05cm);
\draw [fill=black] (-3,-0.5) circle (0.05cm);

\draw [fill=black] (-4,-1) circle (0.05cm);
\draw [fill=black] (-5,0) circle (0.05cm);
\draw [fill=black] (-4.5,-0.5) circle (0.05cm);

% coarse grid right
\draw [line width=0.8pt] (2.5,2.5)-- (2.5,-2.5);
\draw [line width=0.8pt] (4.5,2.5)-- (4.5,-2.5);
\draw [line width=0.8pt] (6.5,2.5)-- (6.5,-2.5);
\draw [line width=0.8pt] (7.5,2.5)-- (7.5,-2.5);

\draw [line width=0.8pt] (2.5,2.5)-- (7.5,2.5);
\draw [line width=0.8pt] (2.5,0.5)-- (7.5,0.5);
\draw [line width=0.8pt] (2.5,-1.5)-- (7.5,-1.5);
\draw [line width=0.8pt] (2.5,-2.5)-- (7.5,-2.5);

% labels right
\draw (3.5,1.8) node[anchor=center] {\tiny{$ \textbf{f}([x_{11}, x_{12}, $}};
\draw (3.7,1.3) node[anchor=center] {\tiny{$ x_{21}, x_{22}]) $}};

\draw (7,1.8) node[anchor=center] {\tiny{$ \textbf{f}([x_{1n},$}};
\draw (7.2,1.3) node[anchor=center] {\tiny{$x_{2n}]) $}};

\draw (3.5,-2) node[anchor=center] {\tiny{$ \textbf{f}([x_{n1}, x_{n2}]) $}};
\draw (7,-2) node[anchor=center] {\tiny{$ \textbf{f}(x_{nm}) $}};
\draw (7,-3) node[anchor=center] {\footnotesize{$n'\times m'$}};

% dots right
\draw [fill=black] (6,1.5) circle (0.05cm);
\draw [fill=black] (5,1.5) circle (0.05cm);
\draw [fill=black] (5.5,1.5) circle (0.05cm);

\draw [fill=black] (6,-2) circle (0.05cm);
\draw [fill=black] (5,-2) circle (0.05cm);
\draw [fill=black] (5.5,-2) circle (0.05cm);

\draw [fill=black] (3.5,0) circle (0.05cm);
\draw [fill=black] (3.5,-1) circle (0.05cm);
\draw [fill=black] (3.5,-0.5) circle (0.05cm);

\draw [fill=black] (7,0) circle (0.05cm);
\draw [fill=black] (7,-1) circle (0.05cm);
\draw [fill=black] (7,-0.5) circle (0.05cm);

\draw [fill=black] (6,-1) circle (0.05cm);
\draw [fill=black] (5,0) circle (0.05cm);
\draw [fill=black] (5.5,-0.5) circle (0.05cm);

% coarsening function
\draw [line width=0.4pt] (1,1)-- (1,-1);
\draw [line width=0.4pt] (-1,0.5)-- (-1,-0.5);
\draw [line width=0.4pt] (-1,0.5)-- (1,1);
\draw [line width=0.4pt] (-1,-0.5)-- (1,-1);
\draw [->,line width=0.4pt] (-2,0) -- (-1,0);
\draw [->,line width=0.4pt] (1,0) -- (2,0);
\draw (0,0) node[anchor=center] {$\Phi_{\beta}$};

% \draw (0,-1.5) node[anchor=center] {\footnotesize restriction};
% \draw (0,-2) node[anchor=center] {\footnotesize operator};
\draw (0,2) node[anchor=center] {\footnotesize $m'= \lfloor \beta m \rfloor $};
\draw (0,1.5) node[anchor=center] {\footnotesize $n'= \lfloor \beta n \rfloor$};

\end{tikzpicture}
        }} \\
        \subfloat[prolongation operator, $\Phi_{\beta}^{-1}$]{\resizebox{0.9\columnwidth}{!} {
        \begin{tikzpicture}[line cap=round,line join=round,>=triangle 45,x=1cm,y=1cm, scale=0.8]
\clip(-8,-4) rectangle (8,3);

% coarse grid left
\draw [line width=0.8pt] (-7.5,2.5)-- (-7.5,-2.5);
\draw [line width=0.8pt] (-5.5,2.5)-- (-5.5,-2.5);
\draw [line width=0.8pt] (-3.5,2.5)-- (-3.5,-2.5);
\draw [line width=0.8pt] (-2.5,2.5)-- (-2.5,-2.5);

\draw [line width=0.8pt] (-7.5,2.5)-- (-2.5,2.5);
\draw [line width=0.8pt] (-7.5,0.5)-- (-2.5,0.5);
\draw [line width=0.8pt] (-7.5,-1.5)-- (-2.5,-1.5);
\draw [line width=0.8pt] (-7.5,-2.5)-- (-2.5,-2.5);

% labels
\draw (-6.5,1.5) node[anchor=center] {\scriptsize{$x_{11}$}};
\draw (-3,1.5) node[anchor=center] {\scriptsize{$x_{1m'}$}};
\draw (-6.5,-2) node[anchor=center] {\scriptsize{$x_{n'1}$}};
\draw (-3,-2) node[anchor=center] {\scriptsize{$x_{n'm'}$}};
\draw (-3,-3) node[anchor=center] {\footnotesize{$n'\times m'$}};

% dots left
\draw [fill=black] (-4,1.5) circle (0.05cm);
\draw [fill=black] (-5,1.5) circle (0.05cm);
\draw [fill=black] (-4.5,1.5) circle (0.05cm);

\draw [fill=black] (-4,-2) circle (0.05cm);
\draw [fill=black] (-5,-2) circle (0.05cm);
\draw [fill=black] (-4.5,-2) circle (0.05cm);

\draw [fill=black] (-6.5,0) circle (0.05cm);
\draw [fill=black] (-6.5,-1) circle (0.05cm);
\draw [fill=black] (-6.5,-0.5) circle (0.05cm);

\draw [fill=black] (-3,0) circle (0.05cm);
\draw [fill=black] (-3,-1) circle (0.05cm);
\draw [fill=black] (-3,-0.5) circle (0.05cm);

\draw [fill=black] (-4,-1) circle (0.05cm);
\draw [fill=black] (-5,0) circle (0.05cm);
\draw [fill=black] (-4.5,-0.5) circle (0.05cm);

% fine grid right
\draw [dashed, line width=0.4pt] (2.5,2.5)-- (2.5,-2.5);
\draw [dashed, line width=0.4pt] (3.5,2.5)-- (3.5,-2.5);
\draw [dashed, line width=0.4pt] (4.5,2.5)-- (4.5,-2.5);
\draw [dashed, line width=0.4pt] (6.5,2.5)-- (6.5,-2.5);
\draw [dashed, line width=0.4pt] (7.5,2.5)-- (7.5,-2.5);

\draw [dashed, line width=0.4pt] (2.5,2.5)-- (7.5,2.5);
\draw [dashed, line width=0.4pt] (2.5,1.5)-- (7.5,1.5);
\draw [dashed, line width=0.4pt] (2.5,0.5)-- (7.5,0.5);
\draw [dashed, line width=0.4pt] (2.5,-1.5)-- (7.5,-1.5);
\draw [dashed, line width=0.4pt] (2.5,-2.5)-- (7.5,-2.5);

% coarse grid right
\draw [line width=0.8pt] (2.5,2.5)-- (2.5,-2.5);
\draw [line width=0.8pt] (4.5,2.5)-- (4.5,-2.5);
\draw [line width=0.8pt] (6.5,2.5)-- (6.5,-2.5);
\draw [line width=0.8pt] (7.5,2.5)-- (7.5,-2.5);

\draw [line width=0.8pt] (2.5,2.5)-- (7.5,2.5);
\draw [line width=0.8pt] (2.5,0.5)-- (7.5,0.5);
\draw [line width=0.8pt] (2.5,-1.5)-- (7.5,-1.5);
\draw [line width=0.8pt] (2.5,-2.5)-- (7.5,-2.5);

% labels right
\draw (3,2) node[anchor=center] {\scriptsize{$x_{11}$}};
\draw (4,2) node[anchor=center] {\scriptsize{$x_{11}$}};
\draw (7,2) node[anchor=center] {\scriptsize{$x_{1m'}$}};

\draw (3,1) node[anchor=center] {\scriptsize{$x_{11}$}};
\draw (4,1) node[anchor=center] {\scriptsize{$x_{11}$}};
\draw (7,1) node[anchor=center] {\scriptsize{$x_{1m'}$}};

\draw (3,-2) node[anchor=center] {\scriptsize{$x_{n'1}$}};
\draw (4,-2) node[anchor=center] {\scriptsize{$x_{n'1}$}};
\draw (7,-2) node[anchor=center] {\scriptsize{$x_{n'm'}$}};

\draw (7,-3) node[anchor=center] {\footnotesize{$n\times m$}};

% dots right
\draw [fill=black] (6,2) circle (0.05cm);
\draw [fill=black] (5,2) circle (0.05cm);
\draw [fill=black] (5.5,2) circle (0.05cm);

\draw [fill=black] (6,1) circle (0.05cm);
\draw [fill=black] (5,1) circle (0.05cm);
\draw [fill=black] (5.5,1) circle (0.05cm);

\draw [fill=black] (6,-2) circle (0.05cm);
\draw [fill=black] (5,-2) circle (0.05cm);
\draw [fill=black] (5.5,-2) circle (0.05cm);

\draw [fill=black] (3,0) circle (0.05cm);
\draw [fill=black] (3,-1) circle (0.05cm);
\draw [fill=black] (3,-0.5) circle (0.05cm);

\draw [fill=black] (4,0) circle (0.05cm);
\draw [fill=black] (4,-1) circle (0.05cm);
\draw [fill=black] (4,-0.5) circle (0.05cm);

\draw [fill=black] (7,0) circle (0.05cm);
\draw [fill=black] (7,-1) circle (0.05cm);
\draw [fill=black] (7,-0.5) circle (0.05cm);

\draw [fill=black] (6,-1) circle (0.05cm);
\draw [fill=black] (5,0) circle (0.05cm);
\draw [fill=black] (5.5,-0.5) circle (0.05cm);

% inverse coarsening function
\draw [line width=0.4pt] (-1,1)-- (-1,-1);
\draw [line width=0.4pt] (1,0.5)-- (1,-0.5);
\draw [line width=0.4pt] (1,0.5)-- (-1,1);
\draw [line width=0.4pt] (1,-0.5)-- (-1,-1);
\draw [->,line width=0.4pt] (-2,0) -- (-1,0);
\draw [->,line width=0.4pt] (1,0) -- (2,0);
\draw (0,0) node[anchor=center] {$\Phi^{-1}_{\beta}$};

% \draw (0,-1.5) node[anchor=center] {\footnotesize prolongation};
% \draw (0,-2) node[anchor=center] {\footnotesize operator};
% \draw (0,2) node[anchor=center] {\footnotesize $m'=\textbf{int}(\beta m)$};
% \draw (0,1.5) node[anchor=center] {\footnotesize $n'=\textbf{int}(\beta n)$};

\end{tikzpicture}
        }}
    \end{tabular}
    \caption{illustration for the restriction operator $\Phi_{\beta}$ and prolongation operator $\Phi_{\beta}^{-1}$ for a parameter $x$}
    \label{fig: operators}
\end{figure}

\begin{figure}[ht]
    \centering
    \resizebox{0.95\columnwidth}{!} {\begin{tikzpicture}[line cap=round,line join=round,>=triangle 45,x=1cm,y=1cm, scale=1.0]
\clip(-6,-3) rectangle (6,3);

% agent box
\draw [line width=0.4pt] (-2,2)-- (-2,1);
\draw [line width=0.4pt] (-2,1)-- (2,1);
\draw [line width=0.4pt] (2,1)-- (2,2);
\draw [line width=0.4pt] (2,2)-- (-2,2);

% coarse environment box
\draw [line width=0.4pt] (-1.5,-0.6)-- (-1.5,-2.5);
\draw [line width=0.4pt] (-1.5,-2.5)-- (1.5,-2.5);
\draw [line width=0.4pt] (1.5,-2.5)-- (1.5,-0.6);
\draw [line width=0.4pt] (1.5,-0.6)-- (-1.5,-0.6);

% environment box
\draw [ line width=0.4pt] (-3,-0.2)-- (-3,-2.9);
\draw [ line width=0.4pt] (-3,-2.9)-- (3.5,-2.9);
\draw [ line width=0.4pt] (3.5,-2.9)-- (3.5,-0.2);
\draw [ line width=0.4pt] (3.5,-0.2)-- (-3,-0.2);

% outline
\draw [->,line width=0.4pt] (-1.5,-0.9) -- (-4.2,-0.9);
\draw [->,line width=0.4pt] (-2.7,-2.2) -- (-4.2,-2.2);
\draw [->,line width=0.4pt] (-1,-2.2) -- (-1.9,-2.2);
\draw [line width=0.4pt] (-4.5,-0.9) -- (-4.2,-0.9);
\draw [line width=0.4pt] (-5.5,-2.2) -- (-4.2,-2.2);
\draw [dashed] (-4.2, -0.4) -- (-4.2, -2.8);
\draw [->,line width=0.4pt] (-4.5,1.2) -- (-2,1.2);
\draw [->,line width=0.4pt] (-5.5,1.8) -- (-2,1.8);
\draw [line width=0.4pt] (-4.5,-0.9)-- (-4.5,1.2);
\draw [line width=0.4pt] (-5.5,-2.2)-- (-5.5,1.8);
\draw [line width=0.4pt] (2,1.5) -- (5,1.5);
\draw [->,line width=0.4pt] (5,-1.5) -- (2.7,-1.5);
\draw [->,line width=0.4pt] (1.9,-1.5) -- (1.5,-1.5);
\draw [line width=0.4pt] (5,1.5)-- (5,-1.5);

% prolongation boxes
\draw [line width=0.4pt] (-2.7,-2)-- (-2.7,-2.4);
\draw [line width=0.4pt] (-2.7,-2.4)-- (-1.9,-2.6);
\draw [line width=0.4pt] (-1.9,-2.6)-- (-1.9,-1.8);
\draw [line width=0.4pt] (-1.9,-1.8)-- (-2.7,-2);
\draw (-2.3,-2.2) node[anchor=center] {$\Phi^{-1}_{\beta}$};

% restriction
\draw [line width=0.4pt] (2.7,-1.3)-- (2.7,-1.7);
\draw [line width=0.4pt] (2.7,-1.7)-- (1.9,-1.9);
\draw [line width=0.4pt] (1.9,-1.9)-- (1.9,-1.1);
\draw [line width=0.4pt] (1.9,-1.1)-- (2.7,-1.3);
\draw (2.3,-1.5) node[anchor=center] {$\Phi_{\beta}$};

% labels
\draw (0,1.5) node[anchor=center] {$\pi(A|S)$};
\draw (0,-2.2) node[anchor=center] {$\mathcal{P}_\beta(S_m, A_m)$};
\draw (0,-0.9) node[anchor=center] {$\mathcal{R}_\beta(S_m, A_m, S_{m+1})$};
\draw (2.1,2.1) node[anchor=north west] {$A_m$};
\draw (-4.1,-1) node[anchor=north west] {$R_{m+1}$};
\draw (-4.1,-2.2) node[anchor=north west] {$S_{m+1}$};
\draw (-3.2,1.2) node[anchor=north west] {$R_m$};
\draw (-3.2,2.4) node[anchor=north west] {$S_m$};
\draw (0,2) node[anchor=south] {$\mathrm{Agent}$};
\draw (0,-0.1) node[anchor=south] {$\mathrm{Environment}, \mathcal{E}_{\beta}$};
\end{tikzpicture}}
    \caption{A typical agent-environment interaction in the proposed multi-grid RL framework}
    \label{fig: rl_framework}
\end{figure}
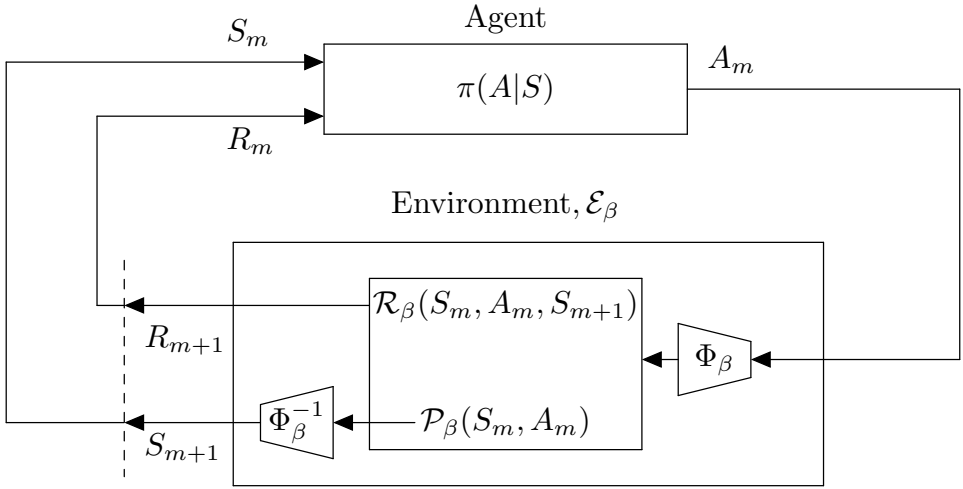

An adaptive multi-grid RL framework is proposed where, essentially, the policies learned using lower grid fidelity environments are transferred and trained with higher fidelity environments. 
The grid fidelity for an environment is described with the factor $\beta \in (0,1]$. 
The environment with $\beta=1$ is assumed to have the fine-grid discretization which guarantees good approximation fluid flow production out of the domain as defined in equation \eqref{eq: obj_fun}.
For any environment where $\beta<1$, the environment grid-size is coarsened with the factor of $\beta$. 
For instance, if a high fidelity environment where $\beta=1$ corresponds to simulation with grid size $64\times64$, the simulation grid size is reduced to $32\times32$ when $\beta$ is set to 0.5. 
Restriction operator $\Phi_{\beta}()$, is used to coarsen the high fidelity simulation parameters with the factor of $\beta$.
This is done by partitioning a finer grid of size $m \times n$ (corresponding to $\beta=1$) into the coarser dimensions $\lfloor \beta m \rfloor \times \lfloor \beta n \rfloor$ (corresponding to $\beta < 1$ where $\lfloor \cdot \rfloor$ is the floor operator) and computing these coarse grid cell values as a function $\textbf{f}$, of values in the corresponding partition.
Figure \ref{fig: operators}a illustrate this restriction operator for a variable  $x \in \mathbb{R}^{n\times m}$.
The function $\textbf{f}$, for different parameters of the reservoir simulation are listed in table \ref{tab:coarse_f}. 
On the other hand, prolongation operator $\Phi^{-1}_{\beta}()$, maps a coarse grid environment parameters to fine grid as shown in figure \ref{fig: operators}b.
\begin{table}
    \caption{ restriction operator function for simulation parameters}
    \centering
    \begin{tabular}{l l }
        \hline
        simulation parameter & function, \textbf{f}  \\
        \hline
        saturation, $s$ & mean \\
        porosity, $\phi$ & mean \\
        pressure, $p$ & mean \\
        permeability, $k$ & harmonic mean \\
        flow control, $a$ & sum \\
        \hline
    \end{tabular}
    \label{tab:coarse_f}
\end{table}
A typical agent-environment interaction using this framework is illustrated in figure \ref{fig: rl_framework}.
Note that the transition function $\mathcal{P}$, and reward function $\mathcal{R}$, are sub-scripted with $\beta$ to indicate the grid fidelity of the environment.
State $S(t_m)$, action $A(t_m)$ and reward $R(t_m)$ are denoted with shorthand notations, $S_m$, $A_m$ and $R_m$, respectively.
Throughout the learning process the policy is represented with states and actions corresponding to high fidelity grid environment. 
As a result, actions and states, to and from the environment, undergo the restriction $\Phi_{\beta}$ and prolongation
$\Phi^{-1}_{\beta}$ operations at each time-step as shown in the environment box of the figure \ref{fig: rl_framework}. 

 \begin{algorithm*}
	\caption{Proximal policy optimisation with adaptive multi-grid framework} 
	\begin{algorithmic}[1]
	\State Define $\delta$, $n$ and an empty array $\textbf{r}$ for convergence criteria
	\State Define a grid fidelity factor array $\boldsymbol{\beta}=[\beta_1, \beta_2,\ldots,\beta_m]$, where $\beta_m=1$ and $\beta_1<\beta_2<\ldots<\beta_m$.
	\State Define an episode limit array $\textbf{E}=[E_1, E_2,\ldots, E_m]$, where $E_1<E_2<\ldots<E_m$.
	\State Define total episode count, $e=0$
	\For{$i=1,2,\ldots, m$}
	    \State Generate the environment $\mathcal{E}_{\beta_i}$, with the grid fidelity factor $\beta_i$
		\For {$iteration=1,2,\ldots$}
			\For {$actor=1,2,\ldots,N$}
				\State Run policy $\pi_{\theta_{old}}$ in environment $\mathcal{E}_{\beta_i}$ , for $T$ time steps (in total, $E$ episodes)
				\State Compute value function estimates $\hat{V}_1,\ldots,\hat{V}_T$ using critic network
				\State Compute advantage function estimates $\hat{A}_1,\ldots,\hat{A}_T$ 
			\EndFor
			\State Optimize $J_{ppo}(\theta)$ with $K$ epochs and minibatch size $M\leq NT$
			\State $\theta_{old}\leftarrow\theta$
			\State Compute the policy return $R^{\pi_{\theta}(A|S)}$ and append it in \textbf{r}
			\State $e:=e+E$
			\If {IsConverged($\textbf{r}$, $n$, $\delta$) or $e \geq E_i$}
			\State \textbf{break}
			\EndIf
		\EndFor
	\EndFor
	\end{algorithmic} 
	\label{alg:ppo_adamultigrid}
\end{algorithm*}

The proposed framework is demonstrated for PPO algorithm. PPO \citep{schulman2017proximal} is a policy gradient algorithm that models the stochastic policy $\pi_\theta(A|S)$, with a neural network (also referred to as the actor network). Essentially, the network parameters $\theta$, are obtained by optimizing for the objective function,
\begin{equation} \label{eq: ppo_loss}
    \begin{split}
        J_{ppo}(\theta) =& \hat{\mathbb{E}}_t  \Bigg[ \Bigg. \min \Big( \Big. r_t(\theta) \hat{Adv} (S_t,A_t),\\  &\textup{clip}(r_t(\theta),1-\epsilon,1+\epsilon)\hat{Adv} (S_t,A_t) \Big. \Big) \Bigg. \Bigg] ,
    \end{split}
\end{equation}
where $r_t(\theta) = \pi_{\theta}(A_t|S_t)/\pi_{\theta_{old}}(A_t|S_t)$ and $\theta_{old}$ correspond to the policy parameters before the policy update.
The advantage function estimator $\hat{Adv}$, is computed using generalized advantage estimator \citep{schulman2015high} which is derived from the value function $V_t$. 
The value function estimator $\hat{V}_t$ is learned through a separate neural network termed as the critic network.
Definitions of advantage and value functions are provided in appendix \ref{app: value_func}.
In practice, a single neural network is used to represent both, actor and critic networks.
The objective function for this integrated actor-critic network is the summation of actor loss term (equation \eqref{eq: ppo_loss}), value loss term and entropy loss term. 
For the purpose of maintaining brevity in our description these latter loss terms are omitted and the policy network's objective function is treated as $J_{ppo}(\theta)$ in further discussion.
However, please note that they are considered while executing the framework.
Readers are referred to \cite{schulman2017proximal} for the detailed definition of policy network loss term.
Algorithm \ref{alg:ppo_adamultigrid} presents the pseudocode for the proposed multi-grid RL framework. 
The framework consists of, in total, $m$ values of grid fidelity factor which are represented with an array $\boldsymbol{\beta}=[\beta_1, \beta_2,\ldots,\beta_m]$, where $\beta_m=1$ and $\beta_1<\beta_2<\ldots<\beta_m$.
The environment is denoted as $\mathcal{E}_{\beta_i}$, which represents the environment with the grid fidelity factor $\beta_i$.
The policy $\pi_{\theta}(A|S)$ is learned initially with environment, $\mathcal{E}_{\beta_1}$, until the convergence criteria is met.
The convergence criteria is checked using the algorithm \ref{alg:isconverged} with predefined parameters $\delta$ and $n$.
Upon convergence, further policy iterations are learned using the environment $\mathcal{E}_{\beta_2}$, and so on until the convergence criteria is met for the highest grid fidelity environment $\mathcal{E}_{\beta_m}$.
A limit for number of episodes to be executed at each grid level is also set.
This is done by defining an episode limit array $\textbf{E}=[E_1, E_2,\ldots, E_m]$, where $E_m$ is total number of episodes to be executed and $E_1<E_2<\ldots<E_m$. 
That is, for every environment with grid fidelity factor $\beta_j$ the maximum number of episodes to be trained is limited to $E_j$.

\section{Case studies}
\label{sect: case studies}
Two test cases are designed representing two distinct permeability uncertainty distributions and control dynamics. 
For both cases, the values for model parameters emulate those in the benchmark reservoir simulation cases, SPE-10 model 2 \citep{christie2001tenth}. 
Table \ref{tab:res_model} delineates these values for test case 1 and 2. 
As per the convention in geostatistics, the distribution of $\log{(k)}$ is assumed to be known and is denoted by $\mathcal{G}$. 
As a result, $g=\log(k)$ is treated as a random variable in the problem description defined in equation \eqref{eq: prob_def}.
Uncertainty distributions for test case 1 and 2 are denoted with $\mathcal{G}_1$ and $\mathcal{G}_2$, respectively.

\begin{table}[ht]
    \caption{Reservoir model parameters}
    \centering
    \begin{tabular}{l l l l}
        \hline
         & case 1 & case 2 & units\\
        \hline
        spatial domain $\mathcal{X}$ & (1200$\times$1200) & (620$\times$1820) & ft$^2$\\
        temporal domain $\mathcal{T}$ & [0,125] & [0,25] & days \\
        initial saturation $s_0$ & 0.0 & 0.0 & -- \\
        viscosity $\mu$ & 0.3 & 0.3 & cP \\
        porosity $\phi$ & 0.2 & 0.2 & -- \\
        number of producers $n_p$ & 31 & 14 & -- \\
        number of injectors $n_i$ & 31 & 7 & -- \\
        total injector flow $\sum a^+$  & 2304 & 9072 & ft$^2$/day\\
        \hline
    \end{tabular}
    \label{tab:res_model}
\end{table}

\subsection{Uncertainty distribution for test case 1}
\begin{figure}
    \centering
    \begin{tabular}{c c}
        \subfloat[test case 1] {\resizebox{0.45\columnwidth}{!} {
        \definecolor{red}{rgb}{1,0,0}
\definecolor{blue}{rgb}{0.08,0.4,0.75}
\begin{tikzpicture}[line cap=round,line join=round,>=triangle 45,x=1cm,y=1cm]
\clip(-3,-3) rectangle (4,3);

% k channel
\fill[line width=0.8pt,fill=black,fill opacity=0.18] (-2,0) -- (2,1) -- (2,0) -- (-2,-1) -- cycle;

% domain
\draw [line width=0.8pt] (2,2)-- (2,-2);
\draw [line width=0.8pt] (2,-2)-- (-2,-2);
\draw [line width=0.8pt] (-2,-2)-- (-2,2);
\draw [line width=0.8pt] (-2,2)-- (2,2);

% extension lines
\draw [line width=0.4pt] (-2,0)-- (-2.6,0);
\draw [line width=0.4pt] (-2,-1)-- (-2.6,-1);
\draw [line width=0.4pt] (-2,2)-- (-2.6,2);
\draw [line width=0.4pt] (2,1)-- (2.6,1);
\draw [line width=0.4pt] (2,2)-- (3.2,2);
\draw [line width=0.4pt] (2,-2)-- (3.2,-2);
\draw [line width=0.4pt] (-2,-2)-- (-2,-2.7);
\draw [line width=0.4pt] (2,-2)-- (2,-2.7);

% dimension lines
\draw [<->,line width=0.4pt] (-2.3,0) -- (-2.3,2);
\draw [<->,line width=0.4pt] (-2.3,0) -- (-2.3,-1);
\draw [<->,line width=0.4pt] (2.3,1) -- (2.3,2);
\draw [<->,line width=0.4pt] (3,-2) -- (3,2);
\draw [<->,line width=0.4pt] (-2,-2.5) -- (2,-2.5);

% dimension label
\draw (-2.3,1) node[anchor=east] {$l_1$};
\draw (-2.3,-0.5) node[anchor=east] {$w$};
\draw (2.3,1.5) node[anchor=west] {$l_2$};
\draw (3,0) node[anchor=west] {$1200$ ft};
\draw (0,-2.5) node[anchor=north] {$1200$ ft};

% wells
\draw [fill=blue] (-1.9,1.9) circle (2.5pt);
\draw [fill=blue] (-1.9,1.7) circle (2.5pt);
\draw [fill=blue] (-1.9,1.5) circle (2.5pt);  
\draw [fill=blue] (-1.9,1.3) circle (2.5pt);
\draw [fill=blue] (-1.9,1.1) circle (2.5pt);
\draw [fill=blue] (-1.9,0.9) circle (2.5pt);
\draw [fill=blue] (-1.9,0.7) circle (2.5pt);
\draw [fill=blue] (-1.9,0.3) circle (2.5pt);
\draw [fill=blue] (-1.9,0.1) circle (2.5pt);
\draw [fill=blue] (-1.9,0.5) circle (2.5pt);
\draw [fill=blue] (-1.9,-1.9) circle (2.5pt);
\draw [fill=blue] (-1.9,-1.7) circle (2.5pt);
\draw [fill=blue] (-1.9,-1.5) circle (2.5pt);
\draw [fill=blue] (-1.9,-1.3) circle (2.5pt);
\draw [fill=blue] (-1.9,-1.1) circle (2.5pt);
\draw [fill=blue] (-1.9,-0.9) circle (2.5pt);
\draw [fill=blue] (-1.9,-0.7) circle (2.5pt);
\draw [fill=blue] (-1.9,-0.3) circle (2.5pt);
\draw [fill=blue] (-1.9,-0.1) circle (2.5pt);
\draw [fill=blue] (-1.9,-0.5) circle (2.5pt);
\draw [fill=red] (1.9,1.9) circle (2.5pt);
\draw [fill=red] (1.9,1.7) circle (2.5pt);
\draw [fill=red] (1.9,1.5) circle (2.5pt);
\draw [fill=red] (1.9,1.3) circle (2.5pt);
\draw [fill=red] (1.9,1.1) circle (2.5pt);
\draw [fill=red] (1.9,0.9) circle (2.5pt);
\draw [fill=red] (1.9,0.7) circle (2.5pt);
\draw [fill=red] (1.9,0.3) circle (2.5pt);
\draw [fill=red] (1.9,0.1) circle (2.5pt);
\draw [fill=red] (1.9,0.5) circle (2.5pt);
\draw [fill=red] (1.9,-1.9) circle (2.5pt);
\draw [fill=red] (1.9,-1.7) circle (2.5pt);
\draw [fill=red] (1.9,-1.5) circle (2.5pt);
\draw [fill=red] (1.9,-1.3) circle (2.5pt);
\draw [fill=red] (1.9,-1.1) circle (2.5pt);
\draw [fill=red] (1.9,-0.9) circle (2.5pt);
\draw [fill=red] (1.9,-0.7) circle (2.5pt);
\draw [fill=red] (1.9,-0.3) circle (2.5pt);
\draw [fill=red] (1.9,-0.1) circle (2.5pt);
\draw [fill=red] (1.9,-0.5) circle (2.5pt);
\end{tikzpicture}
        }}  &
        \subfloat[test case 2] {\resizebox{0.45\columnwidth}{!}{
        \definecolor{red}{rgb}{1,0,0}
\definecolor{blue}{rgb}{0.08,0.4,0.75}
\begin{tikzpicture}[line cap=round,line join=round,>=triangle 45,x=1cm,y=1cm]
\clip(-3,-4) rectangle (3,4);

% domain
\draw [line width=0.8pt] (1.5,3)-- (1.5,-3);
\draw [line width=0.8pt] (1.5,-3)-- (-1.5,-3);
\draw [line width=0.8pt] (-1.5,-3)-- (-1.5,3);
\draw [line width=0.8pt] (-1.5,3)-- (1.5,3);

% extension lines
\draw [line width=0.4pt] (1.5,3) -- (2,3);
\draw [line width=0.4pt] (1.5,-3) -- (2,-3);
\draw [line width=0.4pt] (1.5,-3) -- (1.5,-3.7);
\draw [line width=0.4pt] (-1.5,-3) -- (-1.5,-3.7);

% dimension lines
\draw [<->,line width=0.4pt] (2,3) -- (2,-3);
\draw [<->,line width=0.4pt] (1.5,-3.5) -- (-1.5,-3.5);

% dimension labels
\draw (0,-3.6) node[anchor=north] {$620$ ft};
\draw (2,0) node[anchor=west] {$1820$ ft};

%wells
\draw [fill=red] (-1.4,2.9) circle (2.5pt);
\draw [fill=red] (-1.4,2) circle (2.5pt);
\draw [fill=red] (-1.4,1) circle (2.5pt);
\draw [fill=red] (-1.4,0) circle (2.5pt);
\draw [fill=red] (-1.4,-1) circle (2.5pt);
\draw [fill=red] (-1.4,-2) circle (2.5pt);
\draw [fill=red] (-1.4,-2.9) circle (2.5pt);

\draw [fill=red] (1.4,2.9) circle (2.5pt);
\draw [fill=red] (1.4,2) circle (2.5pt);
\draw [fill=red] (1.4,1) circle (2.5pt);
\draw [fill=red] (1.4,0) circle (2.5pt);
\draw [fill=red] (1.4,-1) circle (2.5pt);
\draw [fill=red] (1.4,-2) circle (2.5pt);
\draw [fill=red] (1.4,-2.9) circle (2.5pt);

\draw [fill=blue] (0,2.9) circle (2.5pt);
\draw [fill=blue] (0,2) circle (2.5pt);
\draw [fill=blue] (0,1) circle (2.5pt);
\draw [fill=blue] (0,0) circle (2.5pt);
\draw [fill=blue] (0,-1) circle (2.5pt);
\draw [fill=blue] (0,-2) circle (2.5pt);
\draw [fill=blue] (0,-2.9) circle (2.5pt);

\end{tikzpicture}
        }}
    \end{tabular}
    \caption{schematic of the spatial domain for test case 1 and 2}
    \label{fig:domain_schema}
\end{figure}
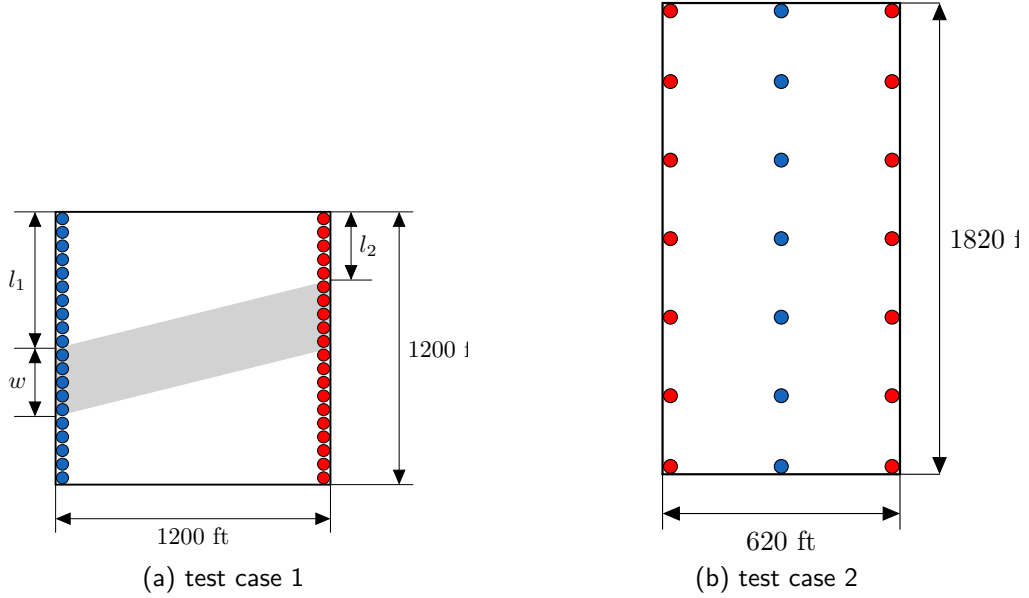

The log-permeability uncertainty distribution for test case 1 is inspired from the case study done by \cite{brouwer2001recovery}.
Figure \ref{fig:domain_schema}a shows schematics of the spatial domain for this case.
In total, 31 injector wells (illustrated with blue circles) and 31 producer wells (illustrated with red circles) are placed at the left and right edge of the domain, respectively.
As illustrated in Figure \ref{fig:domain_schema}a, a linear high permeability channel (shown in gray color) passes from the left to right side of the domain. 
$l_1$ and $l_2$ represent the distance from the top edge of the domain on the left and right side while the channel width is denoted with $w$.
These parameters follow uniform distributions defined as,  $w \sim U(120, 360)$, $l_1 \sim U(0,L-w)$ and $l_2 \sim U(0,L-w)$, where $L$ is domain length.
In other words, the random variable $g$ follows the probability distribution $\mathcal{G}_1$ which is parameterized with $w$, $l_1$ and $l_2$:
\begin{equation*}
    g \sim \mathcal{G}_1(w, l_1, l_2).
\end{equation*}
To be specific, log permeability $g$ at a location $(x,y)$ is formulated as:
\begin{equation*}
    g(x,y) = \left\{\begin{matrix}
\log{(245)} & \textup{if} & \frac{l_2-l_1}{L}x+l_1 \leq y \leq \frac{l_2-l_1}{L}x+l_1+w, \\ 
 &  & \\ 
\log{(0.14)} & \ \ \textup{otherwise}, & 
\end{matrix}\right.
\end{equation*}
where $x$ and $y$ are horizontal and vertical distances from the upper left corner of the domain illustrated in figure \ref{fig:domain_schema}a.
The values for permeability at the channel (245 mD) and the rest of the domain (0.14 mD) are inspired from Upperness log-permeability distribution peak values specified in SPE-10 model 2 case.

% description based on https://web.archive.org/web/20180728153452id_/http://www.cfd.mech.tohoku.ac.jp:80/kawai/publication/selected_papers/AIAA2014-2737_Kawai.jpg
\subsection{Uncertainty distribution for test case 2}
Test case 2 represents uncertainty distribution of a smoother permeability field. 
Figure \ref{fig:domain_schema}b illustrates reservoir domain for this case. 
It comprises of 14 producers (illustrated with red circles) located symmetrically on left and right edges (7 on each edge) of the domain and 7 injectors (illustrated with blue circles) located at the central vertical axis of the domain. 
A prior distribution $F$ is assumed over all the locations $x \in \mathcal{X}$ as,
\begin{align}
    & F(x) = \mu + Z(x), \textup{where}, \label{eq: prior_case2} \\
    & \mathbb{E}(Z(x))=0, \nonumber \\
    & \textup{Cov}(Z(x), Z(\tilde{x})) = \sigma^2 k(x,\tilde{x}), \nonumber
\end{align}
where the process variance, $\sigma$, is assigned as 5 and the exponential covariance function (kernel), $k(x,\tilde{x})$, is defined as,
\begin{equation*}
    k(x,\tilde{x}) = \exp\left [ - \left (  \frac{(x_1 - \tilde{x}_1)^2}{l_1^2} + \frac{(x_2 - \tilde{x}_2)^2}{l_2^2}  \right )^{1/2} \right ],
\end{equation*}
where the parameters $l_1$ and $l_2$ are assigned to be 620ft (width of the domain) and 62ft (10\% of domain width), respectively. 
The posterior distribution given the observed log-permeability vector, $\textbf{g}(x') = [ g(x'_1), g(x'_2), \cdots, g(x'_n)]$, where each observation correspond to a log-permeability value of 2.41 at a well location (i.e., $n=21$ since there are, in total, 21 number of wells in this case). 
From the principle of ordinary kriging, the posterior distribution, $\mathcal{G}_2$, for log-permeability at a location $x \in \mathcal{X}$ is a normal distribution which is defined as,
\begin{align*}
    g(x) \sim &\ \mathcal{G}_2( \hat{g}(x), \hat{s}^2(x)), \textup{ where,}  \\
    \hat{g}(x) =&\ \hat{\mu} + \textbf{k}(x', x)^\intercal \textbf{k}(x',x')^{-1} (\textbf{g}(x')-\textbf{1}\hat{\mu}), \\
    \hat{s}^2(x) =&\ \sigma^2 \Bigg[ \Bigg. 1 - \textbf{k}(x',x)^\intercal \textbf{k}(x',x')^{-1} \textbf{k}(x',x)\\
    &+ \frac{(1-\textbf{1}^\intercal \textbf{k}(x',x')^{-1} \textbf{k}(x',x))^2}{ \textbf{1}^\intercal \textbf{k}(x',x')^{-1} \textbf{1}} \Bigg. \Bigg], 
\end{align*}
where $\textbf{k}(x',x)$ is $n$ dimensional vector whose $i$th value is $k(x'_i, x)$, $\textbf{k}(x',x')$ is $n\times n$ dimensional matrix whose value at $(i,j)$ is $k(x'_i, x'_j)$, $\textbf{1}$ is a $n$ dimensional vector with all elements of one ($\textbf{1} = [1,1,\cdots, 1]^\intercal$) and $\hat{\mu}$ is an estimate of the global mean $\mu$, which is obtained from the kriging model based on the maximum likelihood estimation of the distribution $F(x)$ (from equation \eqref{eq: prior_case2}) for the observations $\textbf{g}(x')$, and is formulated as, 
\begin{equation*}
    \hat{\mu} = \frac{\textbf{1}^\intercal \textbf{k}(x',x')^{-1} \textbf{g}(x')}{\textbf{1}^\intercal \textbf{k}(x',x')^{-1} \textbf{1}}.
\end{equation*}
The log-permeability distribution $\mathcal{G}_2$, is created with an ordinary kriging model using the geostatistics library gstools \citep{sebastian_muller_2019_2541735}. 
In the simulation, samples of the permeability fields are obtained with a clockwise rotation angle of $\pi/8$.

\subsection{State, action and reward formulation} \label{sect: rl_formulation}
PPO algorithm attempts to learn the parameters $\theta$ of the policy neural network $\pi_{\theta}(A|S)$. 
The episodes (i.e. the entire simulation temporal domain $\mathcal{T}$) are divided in five control steps. 
Each episode timestep corresponding to a control step is denoted with $t_m$, where $m \in \{1,2,\cdots, 5 \}$. 
The state $S$, is represented by an observation vector which consists of saturation and pressure values at well locations, $x'$.
Since the saturation values at injector wells are always one,  irrespective of the time $t_m$, they are omitted from the observation vector.
Consequently, the observation vector is of the size $2n_p + n_i$ (i.e., $n_s = 93$ for test case 1 and $n_s = 35$ for test case 2).
Note that this observation vector forms the input to the policy network $\pi_{\theta}(A|S)$.
A vector of flow control values of all the injector and producer wells, denoted by $A$, is represented as the action.
The action vector $A$, consists of in total $n_p + n_i$ values (i.e., $n_a=62$ for test case 1 and $n_a=21$ for test case 2).
In order to maintain constraint defined in equation \eqref{eq: constr}, the  action vector is represented with a vector of weights $w \in \mathbb{R}^{n_a}$, such that $0.001\leq w_j \leq 1$.
Each weight value $w_j$, corresponds to the proportion of flow through the $j$th well.
As a result, the values in the action vector are written as, $( w_1,\cdots,w_{n_i},w_{n_i+1},\cdots,w_{n_i+n_p} )$. 
Flow through $j$th injector $A_j$, is computed such that the constraint defined in \eqref{eq: constr} is satisfied:
\begin{equation*}
    A_j = -\frac{w_j}{\sum_{i=j}^{n_i}w_j}c.
\end{equation*}
Similarly, flow through $j$th producer, $A_{j+n_i}$, is written as,
\begin{equation*}
    A_{j+n_i} = \frac{w_{j+n_i}}{\sum_{j=1}^{n_p}w_{j+n_i}}c.
\end{equation*}
The reward function, as defined in equation \eqref{eq: reward_func}, is divided by total pore volume ($\phi \times lx \times ly $) as form of normalization to obtain a reward function in the range [0,1]. 
The normalized reward represents recovery factor or sweep efficiency of the contaminated fluid.
Recovery factor represents the total amount of contaminants swept out of the domain.
For instance, the recovery factor of 0.65 means that in total of 65\% of contaminants are swept out of the domain using waterflooding. To put it in the context of ground water decontamination problem, the optimal controls correspond to the well controls that maximize the percentage of contaminants swept out of the reservoir.

\subsection{Multi-grid framework formulations}
\begin{figure}
    \centering
    \begin{tabular}{c}
        \subfloat[effect of restriction operator $\Phi_{\beta}$ on a sample of log-permeability]{\includegraphics[width=0.9 \columnwidth]{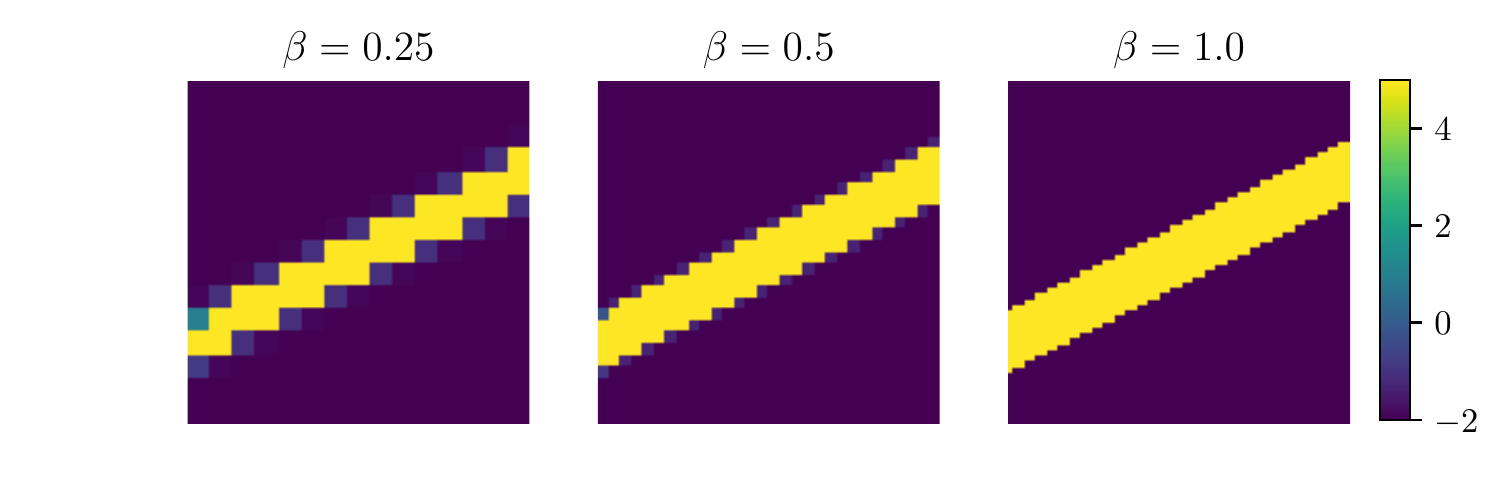}} \\ 
        \subfloat[effect of restriction operator $\Phi_{\beta}$ on saturation corresponding to permeability shown in figure (a)]{\includegraphics[width=0.9\columnwidth]{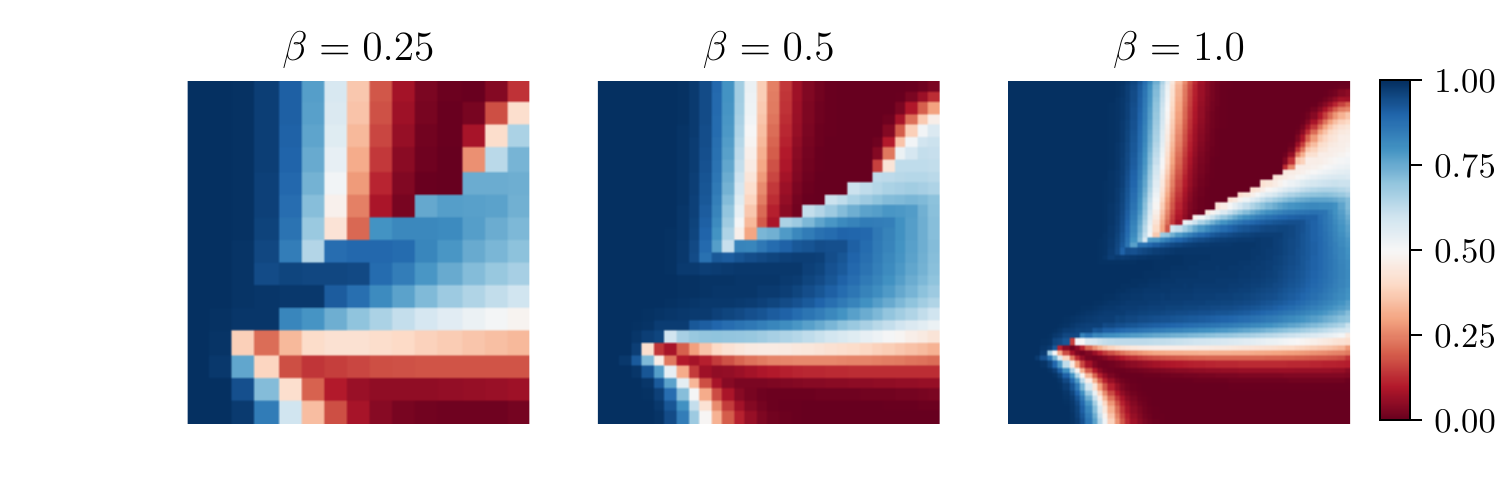}} \\
        \subfloat[effect of grid fidelity on simulation run time]{\includegraphics[width=0.9 \columnwidth]{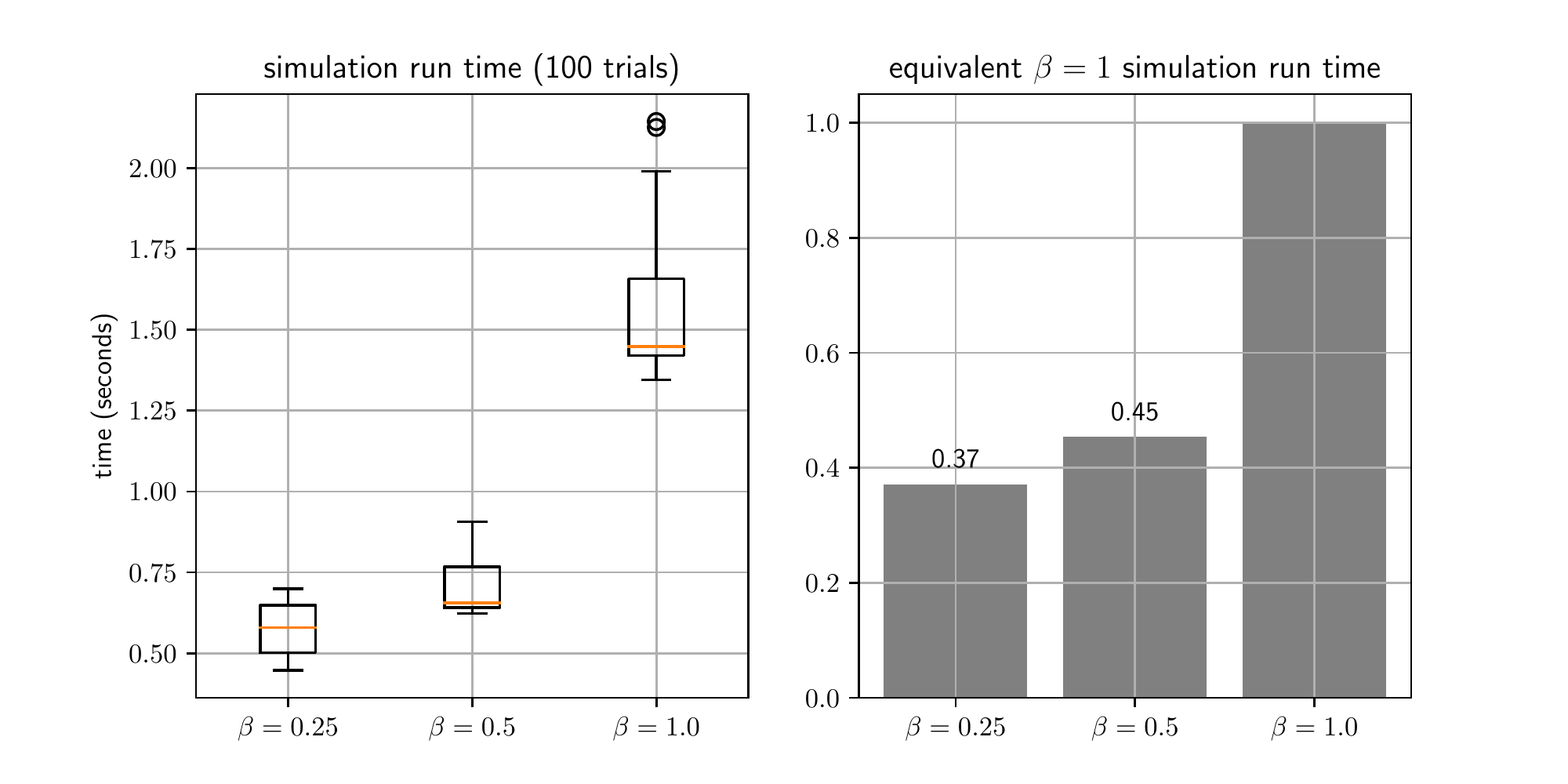}}  
    \end{tabular}
    \caption{effect of grid fidelity factor $\beta$ on the environment for test case 1}
    \label{fig: case_1_coarse}
\end{figure}

\begin{figure}
    \centering
    \begin{tabular}{c}
        \subfloat[effect of restriction operator $\Phi_{\beta}$ on a sample of log-permeability]{\includegraphics[width=0.5 \columnwidth]{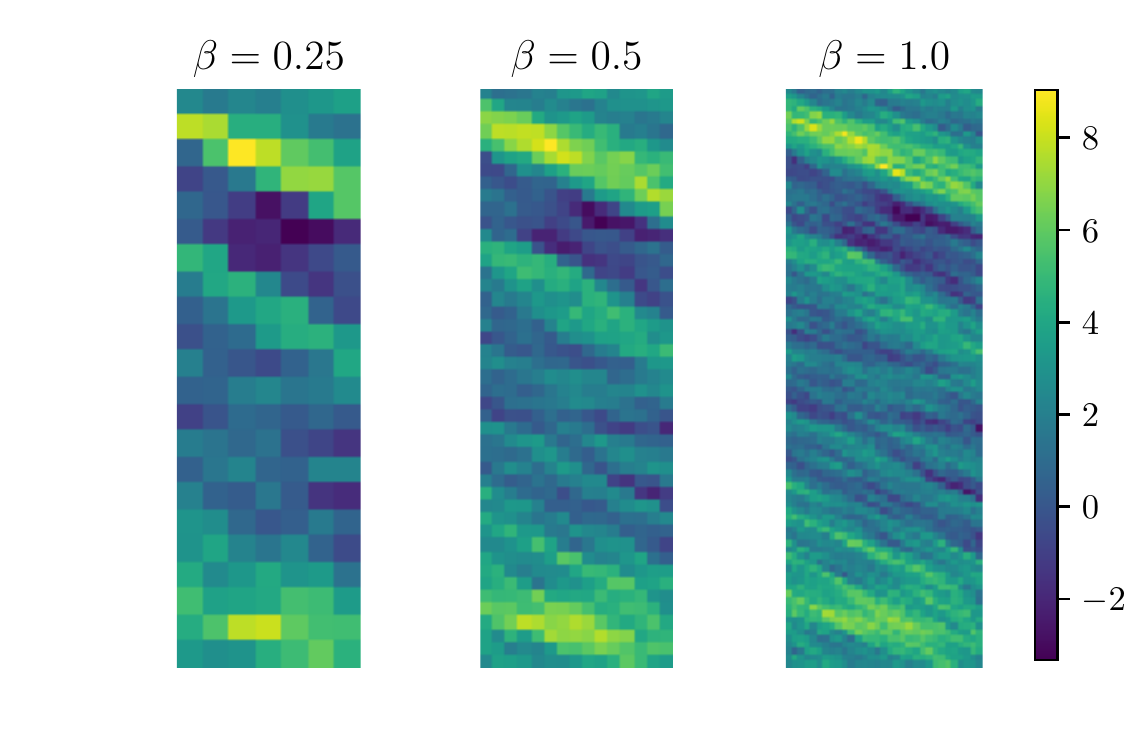}} \\ 
        \subfloat[effect of restriction operator $\Phi_{\beta}$ on saturation corresponding to permeability shown in figure (a)]{\includegraphics[width=0.5 \columnwidth]{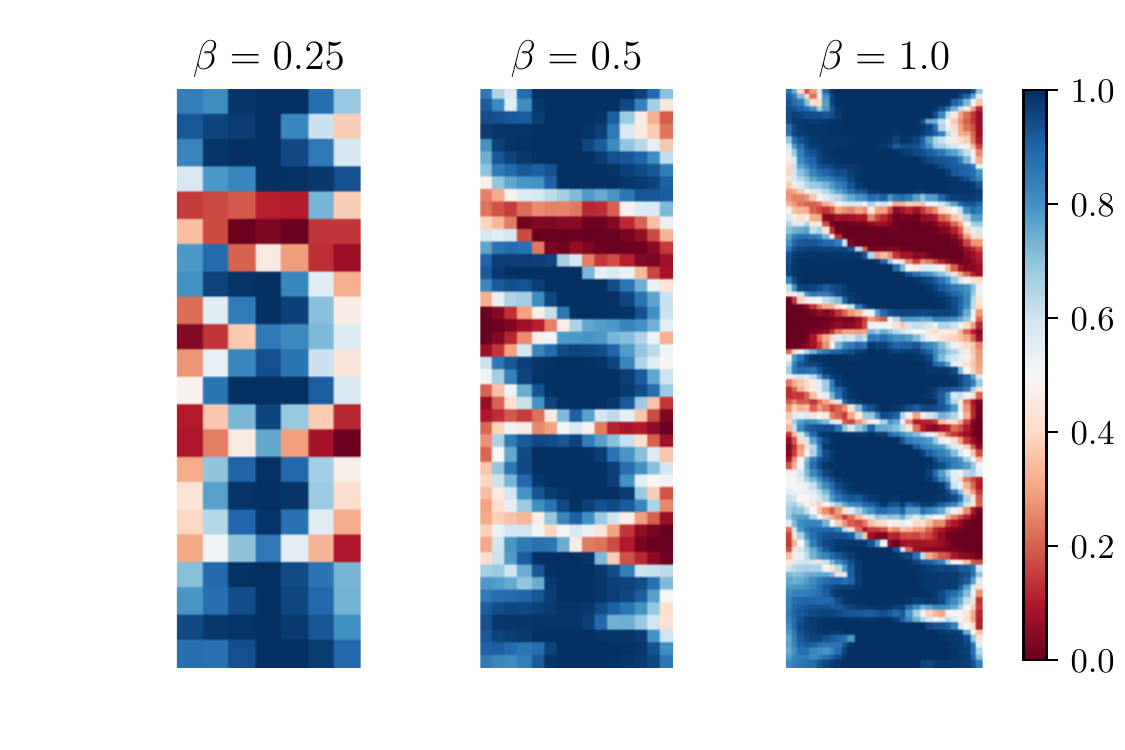}} \\
        \subfloat[effect of grid fidelity on simulation run time]{\includegraphics[width=0.8 \columnwidth]{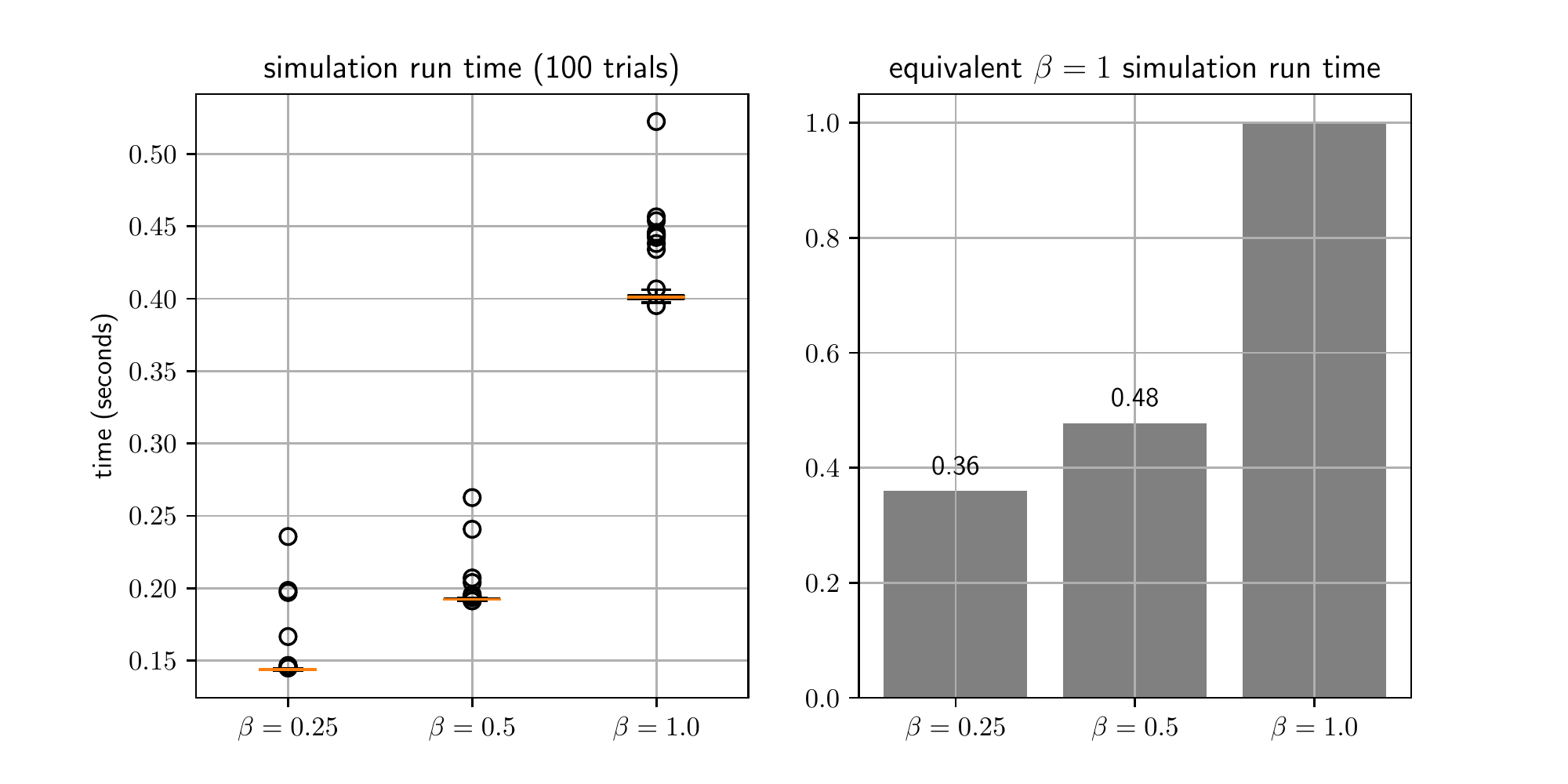}}  
    \end{tabular}
    \caption{effect of grid fidelity factor $\beta$ on the environment for test case 2}
    \label{fig: case_2_coarse}
\end{figure}

The proposed framework is demonstrated using three levels of grid fidelity corresponding to $\beta=0.25$, $\beta=0.5$ and $\beta=1.0$. 
Table \ref{tab: beta_table} lists the discretization grid size corresponding to these grid fidelity factors for both test cases.
In order to show the effectiveness of the proposed framework, the obtained results are compared with single grid and multi-grid frameworks. 
The results for single grid framework are same as if they were obtained using classical PPO algorithm where the environment has a fixed fidelity factor throughout the policy learning process. 
This is done by setting the grid fidelity factor array $\boldsymbol{\beta}$, and episode limit array \textbf{E}, with a single value in algorithm \ref{alg:ppo_adamultigrid}. 
The factor $n$ in convergence criteria procedure (delineated in algorithm \ref{alg:isconverged}) is set to infinity. 
In other words, convergence criteria is unchecked and the policy learning take place for a predefined number of episodes. 
In total three such single-grid experiments are done corresponding to $\beta=0.25$, $\beta=0.5$ and $\beta=1.0$.
Further, two multi-grid experiments are performed to demonstrate the effectiveness of the proposed framework. 
The first multi-grid experiment is referred as ``fixed'' where convergence criteria is kept unchecked just like single-grid frameworks. 
The multiple levels of grids are defined by setting the grid fidelity factor array $\boldsymbol{\beta}$, and episode limit array \textbf{E}, as an array of multiple values corresponding to each fidelity factor value and its corresponding episode count. 
In the fixed multi-grid framework, policy learning takes place by updating the environment fidelity factor according to $\boldsymbol{\beta}$ without checking the convergence criteria (i.e. by setting $n=\infty$). 
Secondly, the ``adaptive'' multi-grid framework parameters are set similar to those used in fixed multi-grid framework except for the convergence criteria parameters $n$ and $\delta$.
Table \ref{tab: exp_type} delineates number of experiments and their corresponding parameters for test case 1 and 2. 
Figure \ref{fig: case_1_coarse} provide visualization of effect of fidelity factor $\beta$, on the simulation of test case 1. 
Figure \ref{fig: case_1_coarse}a and \ref{fig: case_1_coarse}b show log-permeability and saturation plots corresponding to $\beta=0.25$, $\beta=0.5$ and $\beta=1.0$. 
Further, figure \ref{fig: case_1_coarse}c illustrate the effect of grid fidelity on simulation run time for single episode (shown on left with a box plot with 100 simulation run trials) and ``equivalent $\beta=1$ simulation run time'' for each grid fidelity factor (shown on right). 
Equivalent $\beta=1$ simulation run time is defined as the ratio of average simulation run time for a grid fidelity factor $\beta$, to that corresponding to $\beta=1$.
This quantity is used as a scaling factor to convert the number of simulations for any value of $\beta$ to its equivalent number of simulations as if they were performed with $\beta=1$.
Similar plots for test case 2 are demonstrated in figure \ref{fig: case_2_coarse}.

\begin{table}
    \caption{grid fidelity factor and corresponding grid size}
    \centering
    \begin{tabular}{l l l}
        \hline
         & test case 1 & test case 2 \\
        \hline
        $\beta=1$ & $61\times61$ & $31\times91$ \\
        $\beta=0.5$ & $30\times30$ & $15\times45$ \\
        $\beta=0.25$ & $15\times15$ & $7\times22$ \\
        \hline
    \end{tabular}
    \label{tab: beta_table}
\end{table}

Results obtained using the proposed framework are evaluated against the benchmark optimisation results obtained using differential evolution (DE) algorithm \citep{storn1997differential}. For both optimisation methods (PPO and DE) multiprocessing is employed to reduce total computational time. 
However, parallelism behaviour is quite varied between PPO and DE algorithms. 
In PPO algorithms, neural networks are back propagated synchronously at the end of each policy iteration which causes extra computational time in waiting and data distribution. 
As a results, in order to compare computational efforts irrespective of computational resources and parallelism behaviours, it is fair to compare number of simulation runs which is a major source of computational cost in these algorithms. 
The PPO algorithm for the proposed framework is executed using the stable baselines library \citep{stable-baselines3}, while python's SciPy \citep{2020SciPy-NMeth} library is used for DE algorithm. Appendix \ref{app: rl_params} delineate all the algorithm parameters used in this study.

\begin{table*}[t]
    \caption{multi-grid framework experiments}
    \centering
    \begin{tabular}{l l l}
        \hline
         & test case 1 & test case 2 \\
        \hline
         & $\boldsymbol{\beta}=[0.25]$ & $\boldsymbol{\beta}=[0.25]$ \\
        single grid ($\beta=0.25$) & $\textbf{E}=[75000]$ & $\textbf{E}=[150000]$  \\
         & $n=\infty$; $\delta=0$ & $n=\infty$; $\delta=0$ \\
        \hline
         & $\boldsymbol{\beta}=[0.5]$ & $\boldsymbol{\beta}=[0.5]$ \\
        single grid ($\beta=0.5$) & $\textbf{E}=[75000]$ & $\textbf{E}=[150000]$  \\
         & $n=\infty$; $\delta=0$ & $n=\infty$; $\delta=0$ \\
        \hline
         & $\boldsymbol{\beta}=[1.0]$ & $\boldsymbol{\beta}=[1.0]$ \\
        single grid ($\beta=1.0$) & $\textbf{E}=[75000]$ & $\textbf{E}=[150000]$  \\
         & $n=\infty$; $\delta=0$ & $n=\infty$; $\delta=0$ \\
        \hline
        & $\boldsymbol{\beta}=[0.25, 0.5, 1.0]$ & $\boldsymbol{\beta}=[0.25, 0.5, 1.0]$ \\
        fixed multi-grid & $\textbf{E}=[25000, 50000, 75000]$ & $\textbf{E}=[50000, 100000, 150000]$  \\
         & $n=\infty$; $\delta=0$ & $n=\infty$; $\delta=0$ \\
        \hline
        & $\boldsymbol{\beta}=[0.25, 0.5, 1.0]$   & $\boldsymbol{\beta}=[0.25, 0.5, 1.0]$  \\
        adaptive multi-grid & $\textbf{E}=[25000, 50000, 75000]$   & $\textbf{E}=[50000, 100000, 150000]$  \\
        & $n=25$; $\delta=0.2$ & $n=25$; $\delta=0.2$ \\
        \hline
    \end{tabular}
    \label{tab: exp_type}
\end{table*}

\section{Results}
\label{sect: results}
The control policy where injector and producer wells are equally open throughout the entire episode is referred to as the base policy. 
Under such policy, the water flooding prominently takes place in the high permeability region leaving the low permeability region swept inefficiently. 
The optimal policy for these test cases would be to control the producer and injector flow to mitigate this imbalance in water flooding. 
The optimal policy, learned using reinforcement learning for test case 1, show on an average around 12\% improvement with respect to recovery factor achieved using the base policy.
While for test case 2, the average improvement in the order of 25\% is observed.

\begin{figure*}
    \centering
    \includegraphics[width=1.0\textwidth]{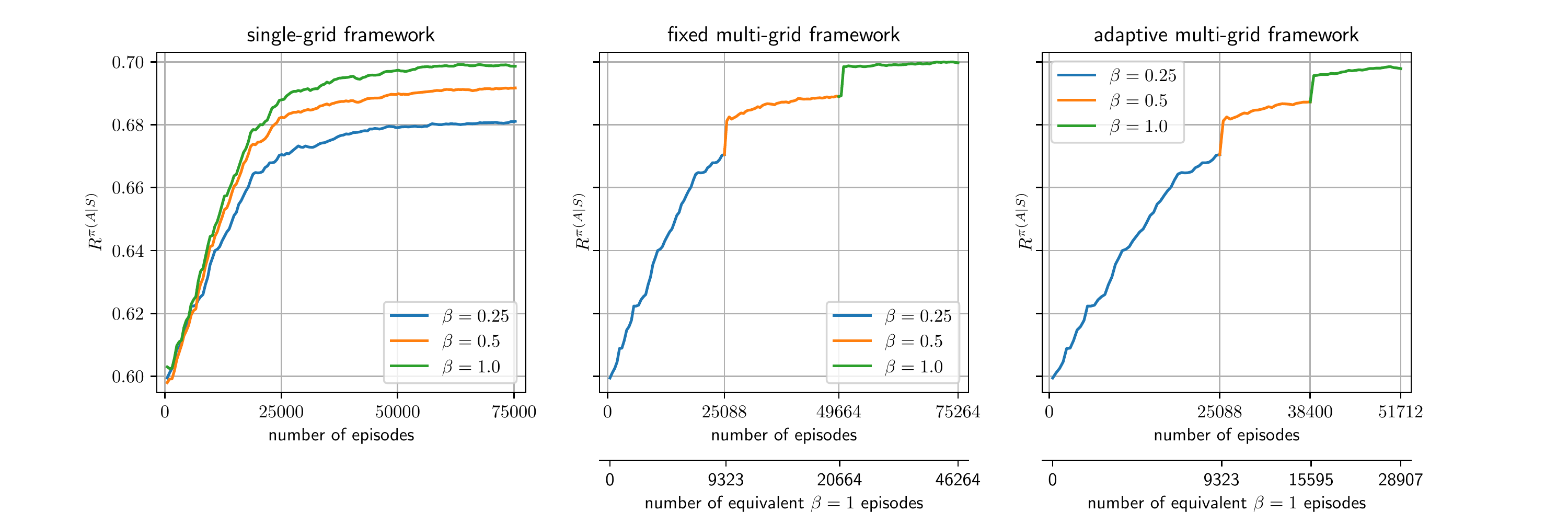}
    \caption{plots of policy return versus number of episodes for test case 1}
    \label{fig: case_1_rl_plots}
\end{figure*}

\begin{figure*}
    \centering
    \begin{tabular}{c}
        \subfloat[samples of log-permeability distribution $\mathcal{G}_1$ used to evaluate the learned policies]{\includegraphics[width=0.7 \textwidth]{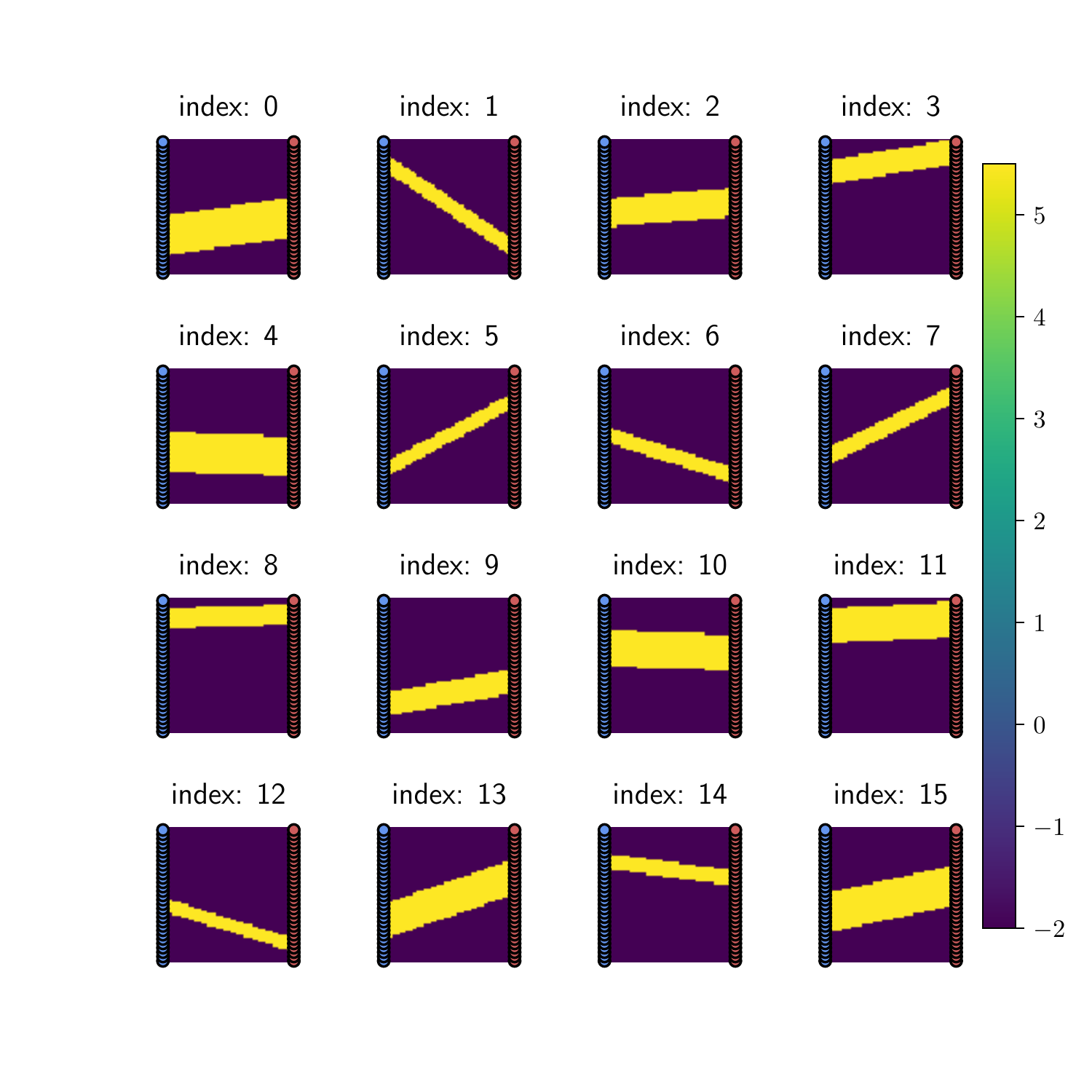}} \\ 
        \subfloat[plot of recovery factor (in \% format) versus evaluation sample index (from figure (a)) for learned policies]{\includegraphics[width=0.8 \textwidth]{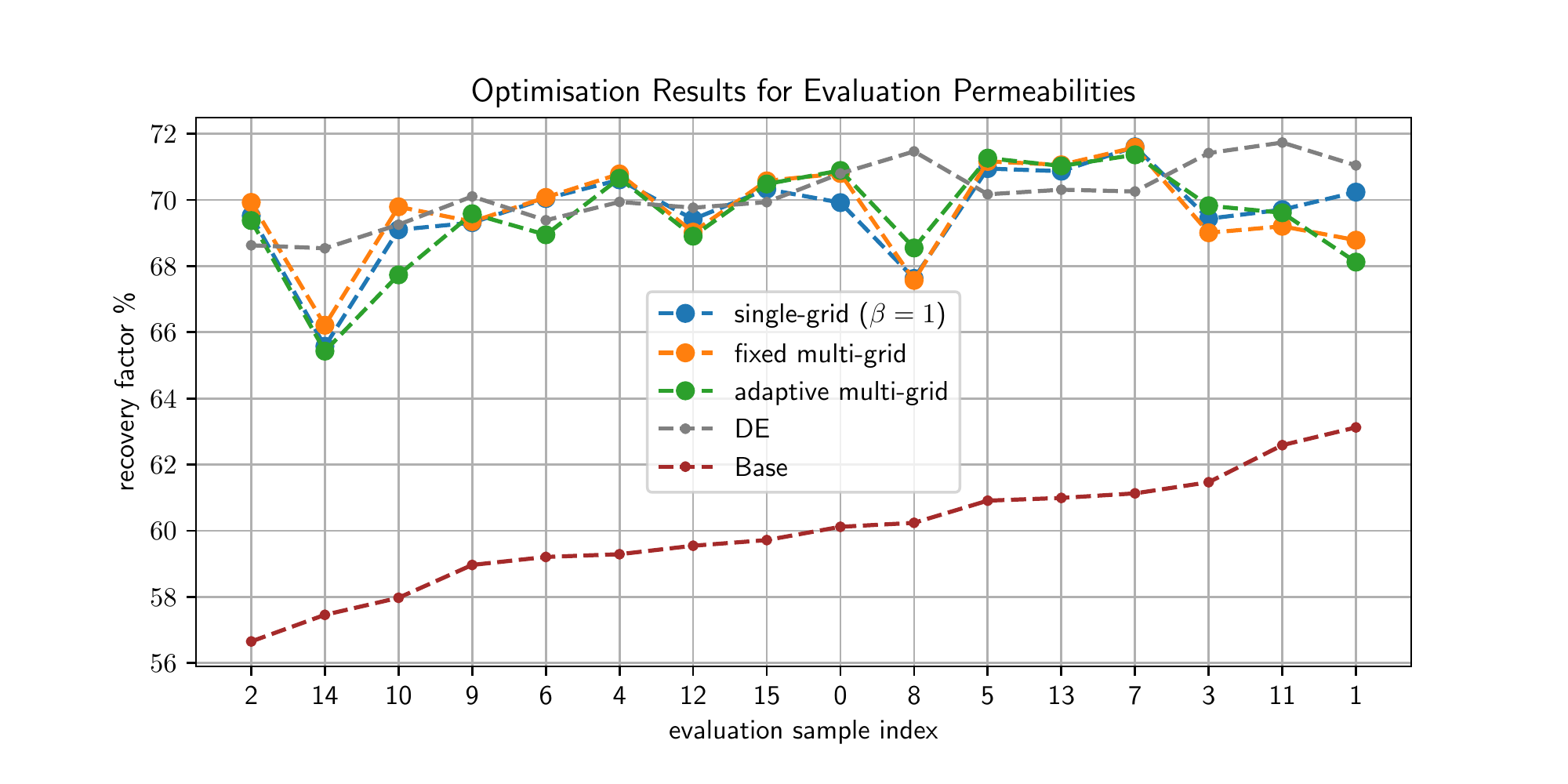}}  
    \end{tabular}
    \caption{evaluation of learned policies for test case 1}
    \label{fig: case_1_results}
\end{figure*}

\begin{figure*}
    \centering
    \includegraphics[width=1.0\textwidth]{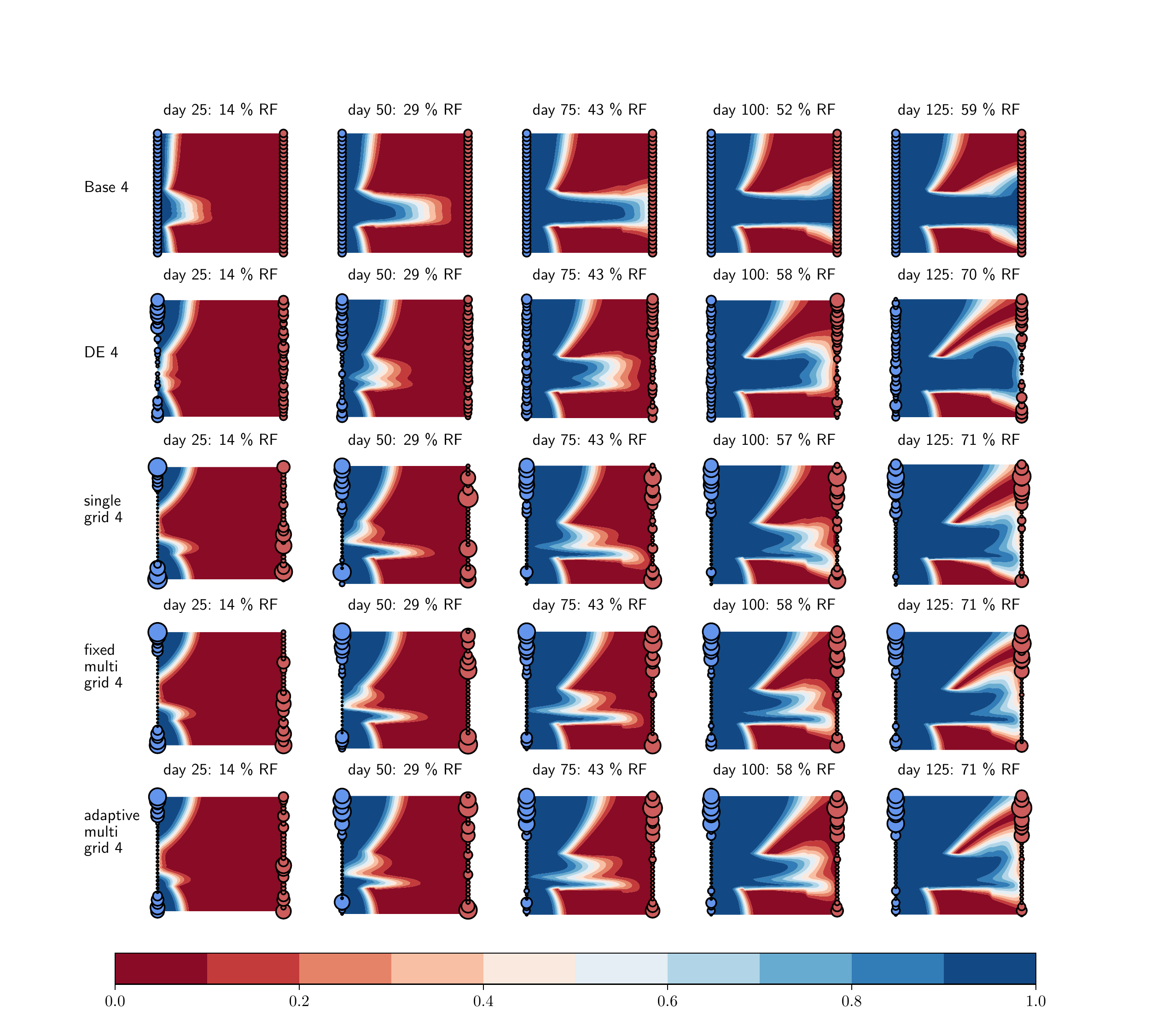}
    \caption{illustration of learned optimal control policies for test case 1}
    \label{fig:case_1_rl_ex}
\end{figure*}

Figure \ref{fig: case_1_rl_plots} illustrates the plots for policy return $R^{\pi(A|S)}$, corresponding to all the frameworks listed in table \ref{tab: exp_type} for test case 1. 
At the beginning of the learning process, the policy return values for single-grid framework keeps improving and eventually converge to a maximum value when the policy converges to an optimal policy.
Note that for lower value of grid fidelity factor $\beta$, the optimal policy return is also low. 
In other words, the coarsening of simulation grid discretization also reflects in overall reduction in recovery factor. 
This is due to the low accuracy of states and actions representation for environments with $\beta<1$.
On the other hand, the overall computational gain is observed due to coarser grid sizes. 
Simulation run time corresponding to $\beta=0.25$ and $\beta=0.5$ show around 66\% and 54\% reduction as compared to that with $\beta=1$. 
the results of multi-grid frameworks are compared with the single grid framework corresponding to $\beta=1$ which refers to classical PPO algorithm using the environment with a fixed high fidelity grid factor.
As shown in the plots at the center and right of figure \ref{fig: case_1_rl_plots}, both multi-grid frameworks show convergence to the optimal policy which is achieved using high fidelity single grid framework.
In the fixed multi-grid framework the fidelity factor, is incremented at a fixed interval of 25000 number of episodes.
The adaptive framework is also provided with the same interval but with additional convergence check within each interval. 
For multi-grid learning plots shown in figure \ref{fig: case_1_rl_plots} (center and right plots), equivalent number of episodes corresponding to the environment with $\beta=1$ is illustrated as a secondary horizontal axis.
This way, the computational effect of multi-grid frameworks is directly compared to single-grid (with $\beta=1$) framework.
The equivalent number of $\beta=1$ episodes corresponding to episodes with certain $\beta$ value are computed by multiplying it with the equivalent $\beta=1$ simulation run time.
For instance, number of episodes with $\beta=0.25$ are multiplied with 0.37. 
For fixed multi-grid framework, it takes 46264 number of equivalent $\beta=1$ episodes to achieve an equally optimal policy that is obtained with 75000 number of episodes using single grid ($\beta=1$) framework.
Similarly, the same is achieved with just 28907 number of equivalent $\beta=1$ episodes using adaptive multi-grid framework.
In other words, around 38\% and 61\% reduction is observed in simulation run time using fixed and adaptive multi-grid frameworks, respectively.
Further, the robustness of the policy learned using these frameworks is compared by applying it on a highest fidelity environment with random permeability samples from the distribution $\mathcal{G}_1$, which were never seen during the policy learning process. 
Figure \ref{fig: case_1_results}a shows the plots of these unseen permeability fields, while the corresponding results obtained using these frameworks are plotted in figure \ref{fig: case_1_results}b.
Optimal results obtained using differential evolutionary (DE) algorithms are provided as benchmark (marked as DE in figure \ref{fig: case_1_results}b).
Note that DE algorithm, in itself, is not a suitable method to solve the robust optimal control problem since it can provide optimal controls only for certain permeability samples as opposed to PPO algorithm where the learned policy is applicable to all samples of permeability distribution.
However, DE results are used as the reference optimal results which are achieved by direct optimization on sample by sample basis.
Equivalence in the optimality of learned policies obtained using these three experiments can be observed from the closeness in their corresponding optimal recovery factors.
Figure \ref{fig:case_1_rl_ex} demonstrate the policy visualization for an example of permeability sample in case 1. 
In this figure, the results are shown for permeability sample index 4 from the figure \ref{fig: case_1_results}a where a high permeability channel passes through lower region of the domain.
The optimal policy, in this case, would be to restrict the flow through injector wells and producer wells which are in the vicinity of the channel.
The super-positioned comparison of optimal results for base case, differential evolution, single-grid framework (where $\beta=1$), fixed multi-grid framework and adaptive multi-grid framework shows that the optimal policy is learned successfully using the proposed framework.

\begin{figure*}
    \centering
    \includegraphics[width=1.0\textwidth]{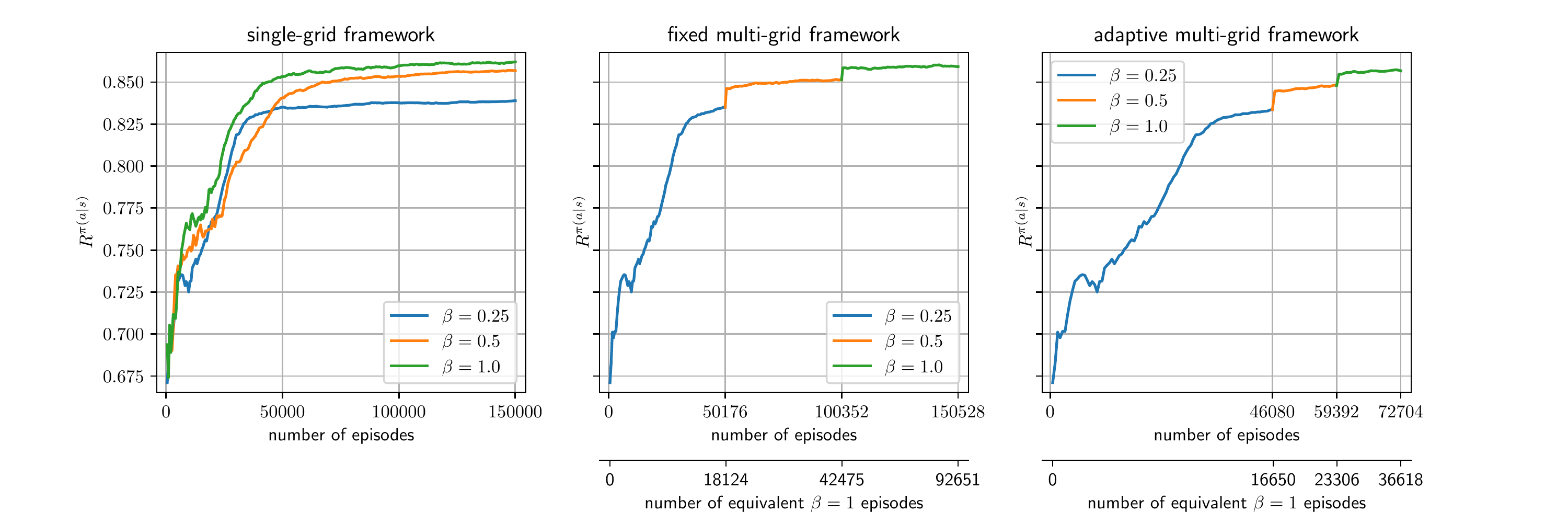}
    \caption{plots of policy return versus number of episodes for test case 2}
    \label{fig: case_2_rl_plots}
\end{figure*}

\begin{figure*}
    \centering
    \begin{tabular}{c}
        \subfloat[samples of log-permeability distribution $\mathcal{G}_2$ used to evaluate the learned policies]{\includegraphics[width=0.7 \textwidth]{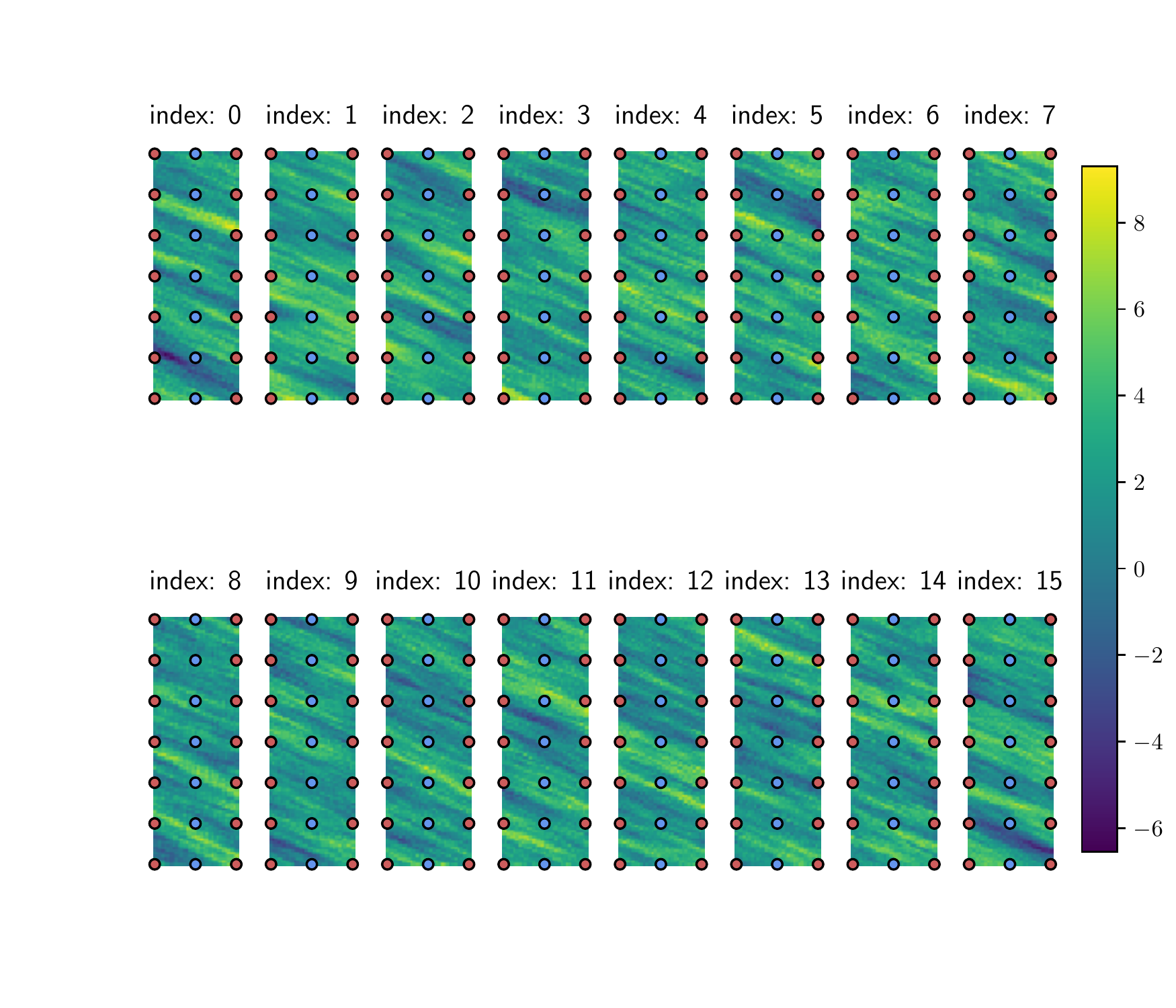}} \\ 
        \subfloat[plot of recovery factor (in \% format) versus evaluation sample index (from figure (a)) for learned policies]{\includegraphics[width=0.8 \textwidth]{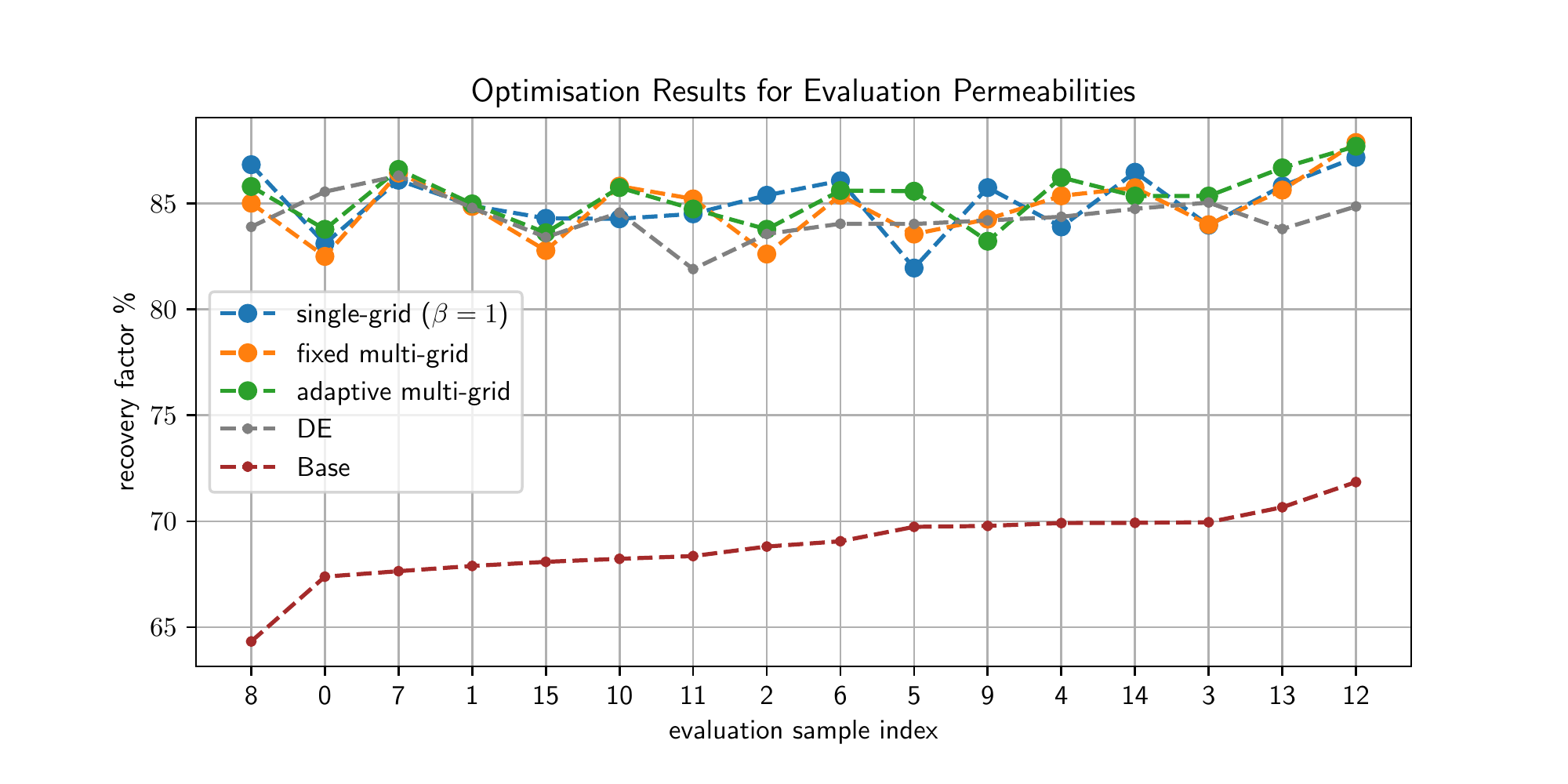}}  
    \end{tabular}
    \caption{evaluation of learned policies for test case 2}
    \label{fig: case_2_results}
\end{figure*}

\begin{figure*}
    \centering
    \includegraphics[trim=10 0 50 0,clip, width=0.6\textwidth, height=1.2\textwidth]{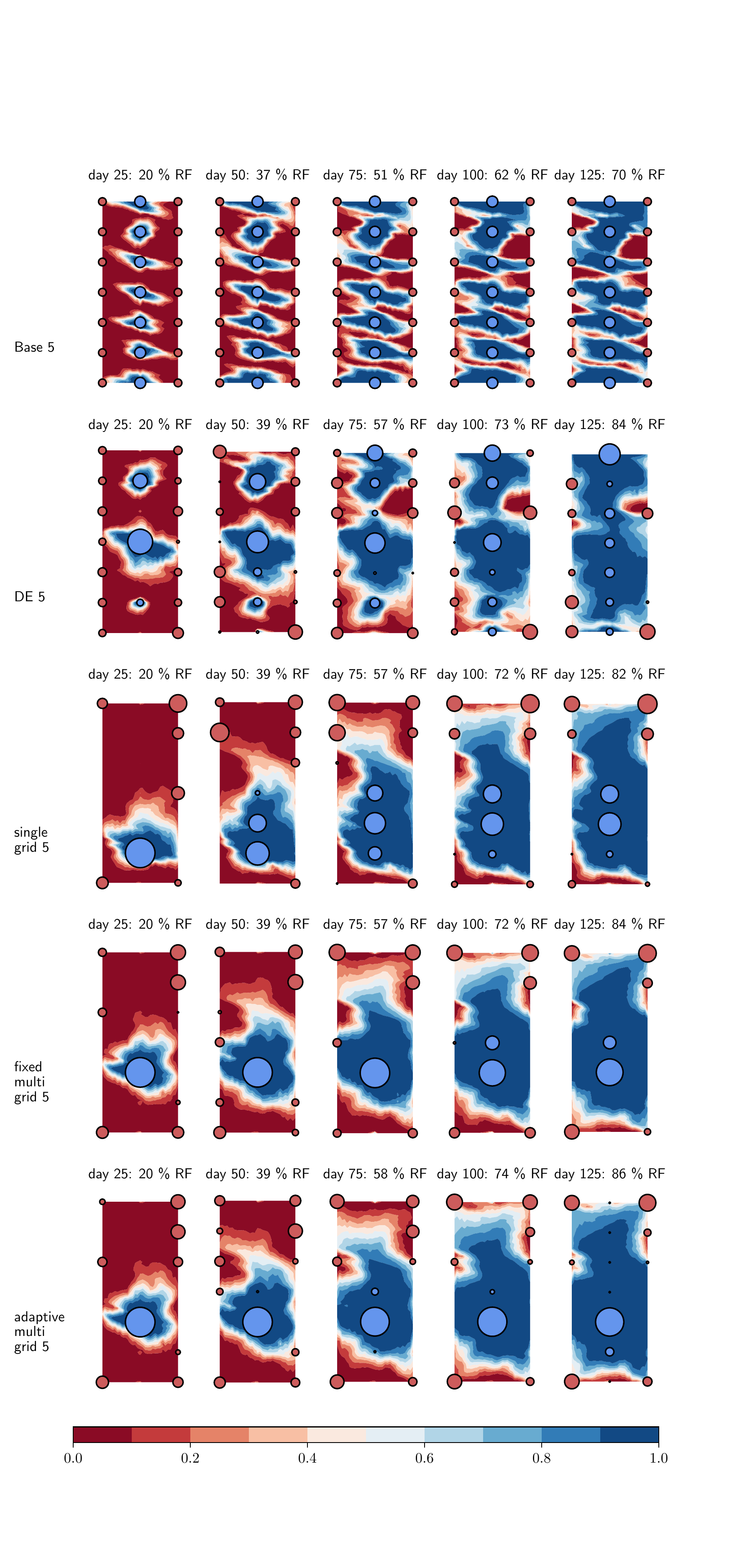}
    \caption{illustration of learned optimal control policies for test case 2}
    \label{fig:case_2_rl_ex}
\end{figure*}

For test case 2, similar results are observed as shown in figure \ref{fig: case_2_rl_plots}. 
The single-grid algorithms converge to an optimal policy in total 150000 number of episodes. 
The fixed multi-grid algorithm is trained with 50000 episode interval for each grid fidelity factor as shown in the central plot in figure \ref{fig: case_2_rl_plots}.
The optimal policy is learned in 92657 equivalent $\beta=1$ episodes thus saving around 38\% of simulation run time.
The adaptive multi-grid framework further reduces computational cost by achieving the optimal policy in 36618 number of equivalent $\beta=1$ episodes (simulation time reduction of about 76\% with respect to $\beta=1$ single-grid framework). 
Figure \ref{fig: case_2_results} illustrate the results of policy evaluation on an unseen permeability samples from the distribution $\mathcal{G}_2$.
The permeability samples are shown in figure \ref{fig: case_2_results}a and the optimal recovery factor corresponding to learned policies are plotted in figure \ref{fig: case_2_results}b. Figure \ref{fig:case_2_rl_ex} demonstrate the optimal controls for an example of permeability sample index 5 from figure \ref{fig: case_2_results}a. 
The optimal policy learned using differential evolution algorithm refers to increasing the flow through injector wells which are in low permeability region while restricting the flow through producer wells for which the water cutoff is reached. 
The policies learned using RL framework takes advantage of the default location and orientation of high permeability regions.
In this case, the optimal policy is achieved by controlling the well flow control such that the flow traverses through the permeability channels (that is, the flow is more or less perpendicular to the permeability orientation).

\section{Conclusion}
\label{sect: conclusion}
An adaptive multi-grid RL framework is introduced to solve robust optimal well control problem. The proposed framework results in significant reduction in computational cost of policy learning process as compared to classical PPO algorithm results. In the presented case studies, 61\% computational savings in simulation runtime for test case 1 and 76\% for test case 2 is observed. The results are highly dependent on the right choice of the algorithm hyper-parameters (e.g. $\delta$, $n$, $\boldsymbol{\beta}$ and $\textbf{E}$) which were tuned heuristically.
As a future direction to this research study, the aim is to find the optimal values for $\boldsymbol{\beta}$ that maximizes the overall computational savings.
Furthermore, policy transfer was performed sequentially in the current framework which seemed to have worked optimally. However, to improve the generality of the proposed framework it would be important to study the effect of sequence of policy transfer on the overall performance.

\section*{Acknowledgment}
The first author would like to acknowledge the Ali Danesh scholarship to fund his PhD studies at Heriot-Watt University. The authors would also like to acknowledge the EPSRC funding through the grant EP/V048899/1.

\clearpage

\appendix

\section{Cluster analysis of permeability uncertainty distribution} \label{app: cluster}
training vector \textbf{k}, is chosen to represent the variability in the permeability distribution $\mathcal{K}$. 
For the optimal control problem, our main interest is in the uncertainty in the dynamical response of permeability rather than the uncertainty in permeability itself. 
As a result, connectivity distance \citep{park2011modeling} is used as a measure of distance between permeability field samples. 
The connectivity distance matrix $\textbf{D} \in \mathbb{R}^{N \times N}$ among $N$ samples of $\mathcal{K}$ is formulated as,
\begin{equation*}
    \textbf{D}(k_i, k_j) = \sum_{x''} \int_{t_0}^{T} \left [ s(x'',t; k_i) - s(x'',t; k_j) \right ]^2 dt,
    \label{eq: conn_dist}
\end{equation*}
where $N$ correspond to a large number number of samples of uncertainty distribution, $s(x'',t;k_i)$ is saturation at location $x''$, and time $t$, when the permeability is set to $k_i$ and all wells are open equally. 
Multi-dimensional scaling of the distance matrix \textbf{D} is used to produce $N$ two-dimensional coordinates $d_1, d_2, \cdots, d_N$, each representing a permeability sample. 
The coordinates $d_1, d_2, \cdots, d_N$ are obtained such that the distance between $d_i$ and $d_j$ is equivalent to $\textbf{D}(k_i, k_j)$. In the k-means clustering process, these coordinates are divided in $l$ sets $S_1, S_2, \cdots , S_l$, obtained by solving the optimisation problem:
\begin{equation*}
    \arg \min_{S} \sum_{i}^{l} \sum_{d_j \in S_i } \left \|   d_j - \mu_{S_i} \right \|,
\end{equation*}
where $\mu_{S_i}$ is average of all coordinates in the set $S_i$. The training vector \textbf{k} is a set of $l$ samples of $\mathcal{K}$ where each of its value $k_i$ correspond to the one nearest to $\mu_{S_i}$.
\begin{figure*}
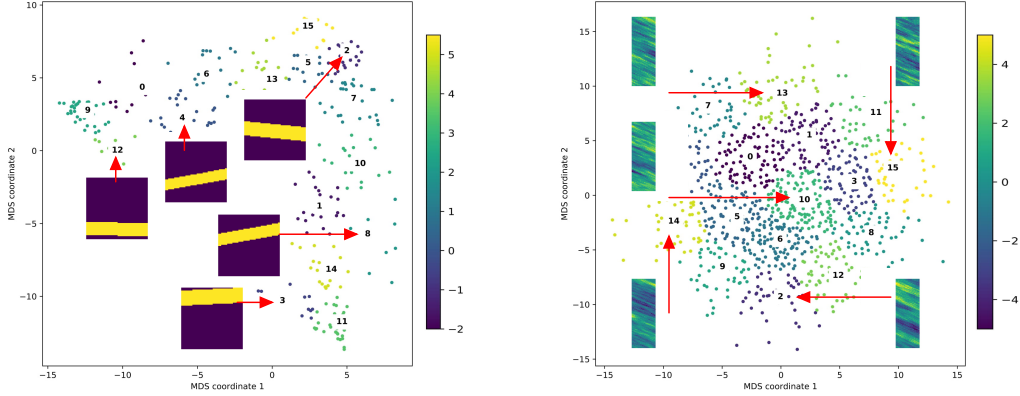

    \centering
    \begin{tabular}{cc}
    
        \subfloat[clustering of $\mathcal{G}_1$ distribution samples] {\resizebox{0.45\textwidth}{!} {
        \input{images/case_1_cluster_image.tikz}
        }}  &
        \subfloat[clustering of $\mathcal{G}_2$ distribution samples] {\resizebox{0.45\textwidth}{!}{
        \input{images/case_2_cluster_image.tikz}
        }}
    \end{tabular}
    \caption{log-permeability plots for training data of test case 1 and 2}
    \label{fig: cluster}
\end{figure*}
Total number of samples $N$ and clusters $l$ are chosen to be 1000 and 16 for both uncertainty distributions, $\mathcal{G}_1$ and $\mathcal{G}_2$. 
Training vector $\textbf{k}$ is obtained with samples $k_1, \cdots, k_{16}$ each corresponding to a cluster center.
Figure \ref{fig: cluster}a and \ref{fig: cluster}b show cluster plots for $\mathcal{G}_1$ and $\mathcal{G}_2$ permeability distribution samples, respectively. Further, 16 permeability samples, each randomly chosen from a cluster, are chosen to evaluate the learned policies. Figures \ref{fig: case_1_results}a and \ref{fig: case_2_results}a illustrate these samples for test case 1 and 2, respectively.

\section{Definitions of value and advantage function} \label{app: value_func}
In RL, the policy $\pi(A|S)$ is said to be optimal if it maps the state $S_t$ with an action $A_t$ that correspond to maximum expected return value. These return values are learned through the data obtained in agent-environment interactions. Following are some definition of return values typically used in RL:

Value function is the expected future return for a particular state $S_t$ and is defined as,
\begin{equation*}
    V(S) = \mathbb{E}_{\pi} \left [ \sum_{m} \gamma^m R_{m+t+1} | S_t = S \right],
\end{equation*}
where $\mathbb{E}_{\pi}[\cdots]$ denotes expected value given that the agent follows the policy $\pi$. As a short hand notation, $V(S)$ at state $S_t$ is denoted as $V_t$.

Q function is similar to value function except that it represent the expected return when the agent takes action $a_t$ in the state $S_t$. It is defined as,
\begin{equation*}
    Q(S, A) = \mathbb{E}_{\pi} \left [ \sum_{m} \gamma^m R_{m+t+1} | S_t = S, A_t = A \right].
\end{equation*}
Advantage function is defined as the difference between Q function and value function and is denoted by $Adv(S, A)$ at state $S$ and action $A$.

\section{Algorithm parameters} \label{app: rl_params}
\begin{figure*}
    \centering
    \begin{tabular}{c}
        \subfloat[seed 1]{\includegraphics[width= \textwidth]{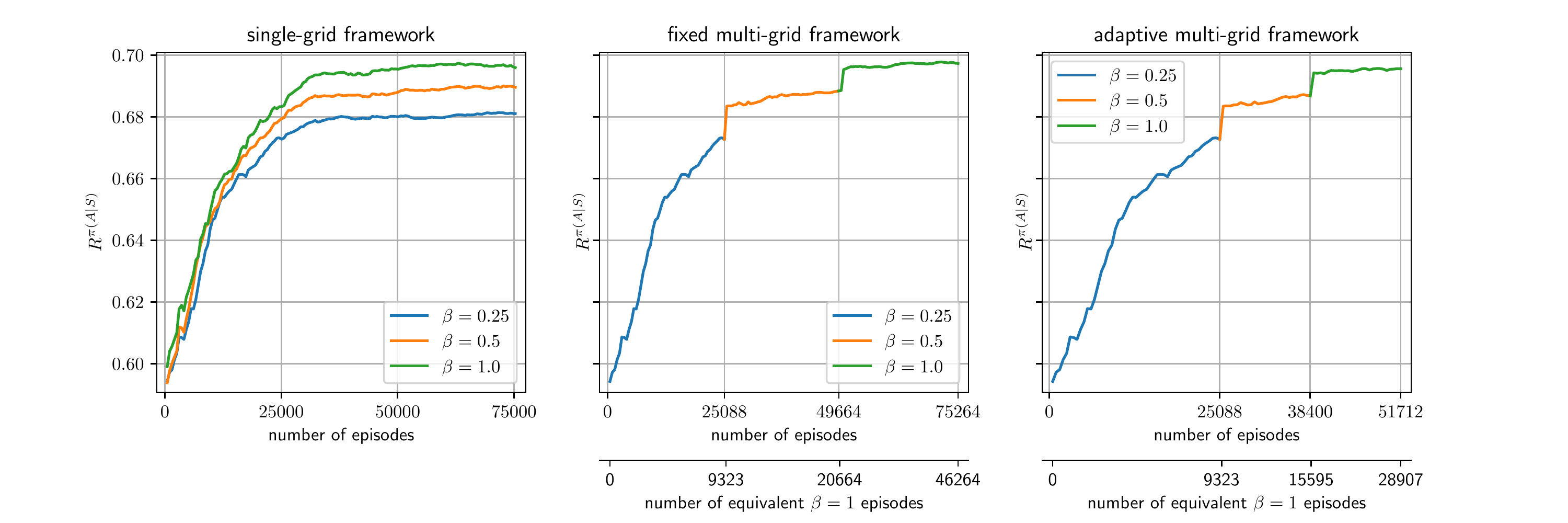}} \\ 
        \subfloat[seed 2]{\includegraphics[width= \textwidth]{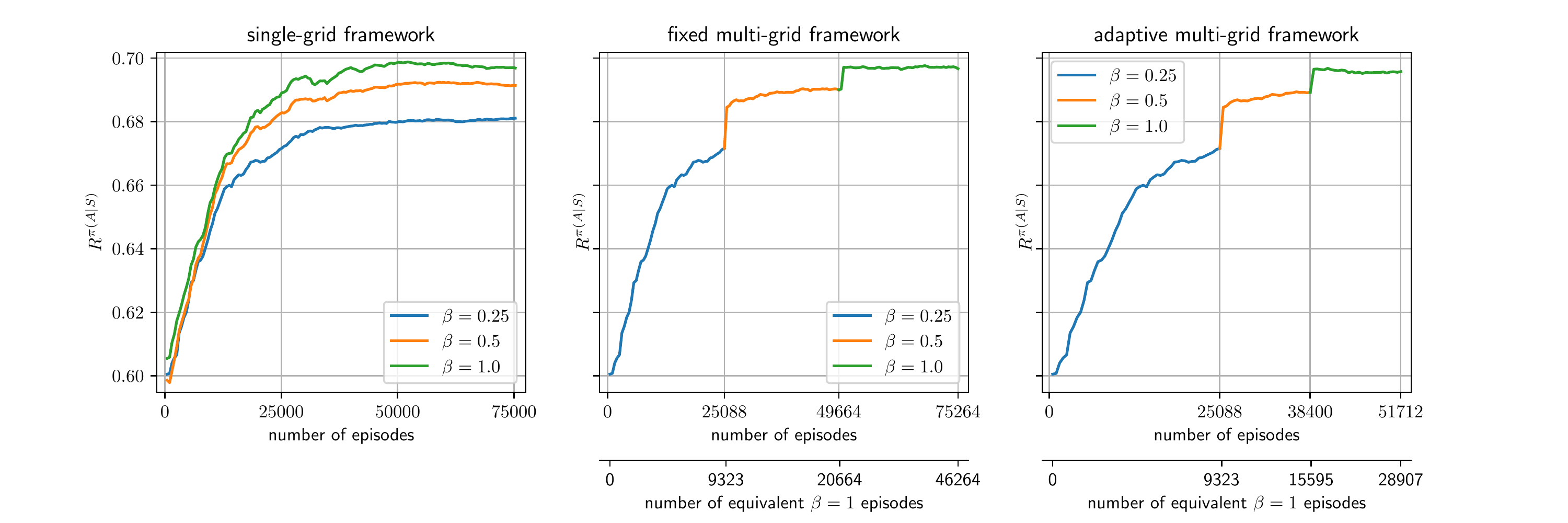}} \\ 
        \subfloat[seed 3]{\includegraphics[width= \textwidth]{images/case_1_seed_3_rl_mg_plots.pdf}} \\ 
    \end{tabular}
    \caption{learning plots for three distinct seed values for test case 1}
    \label{fig: case_1_seeds}
\end{figure*}

\begin{figure*}
    \centering
    \begin{tabular}{c}
        \subfloat[seed 1]{\includegraphics[width= \textwidth]{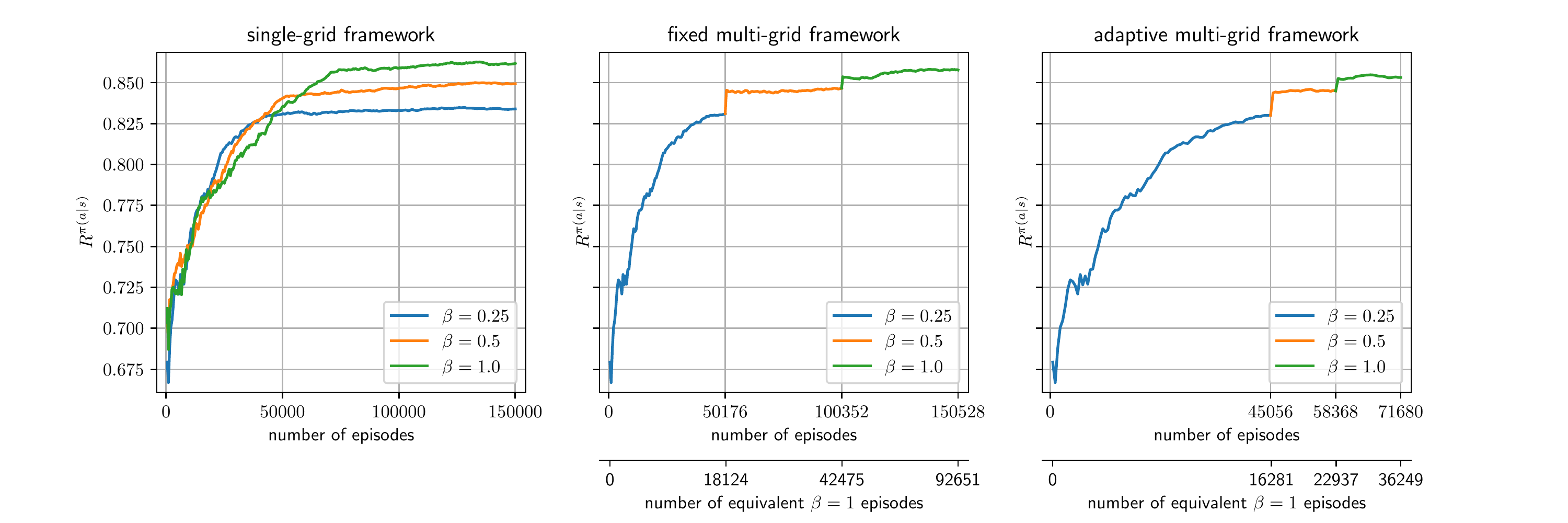}} \\ 
        \subfloat[seed 2]{\includegraphics[width= \textwidth]{images/case_2_seed_2_rl_mg_plots.pdf}} \\ 
        \subfloat[seed 3]{\includegraphics[width= \textwidth]{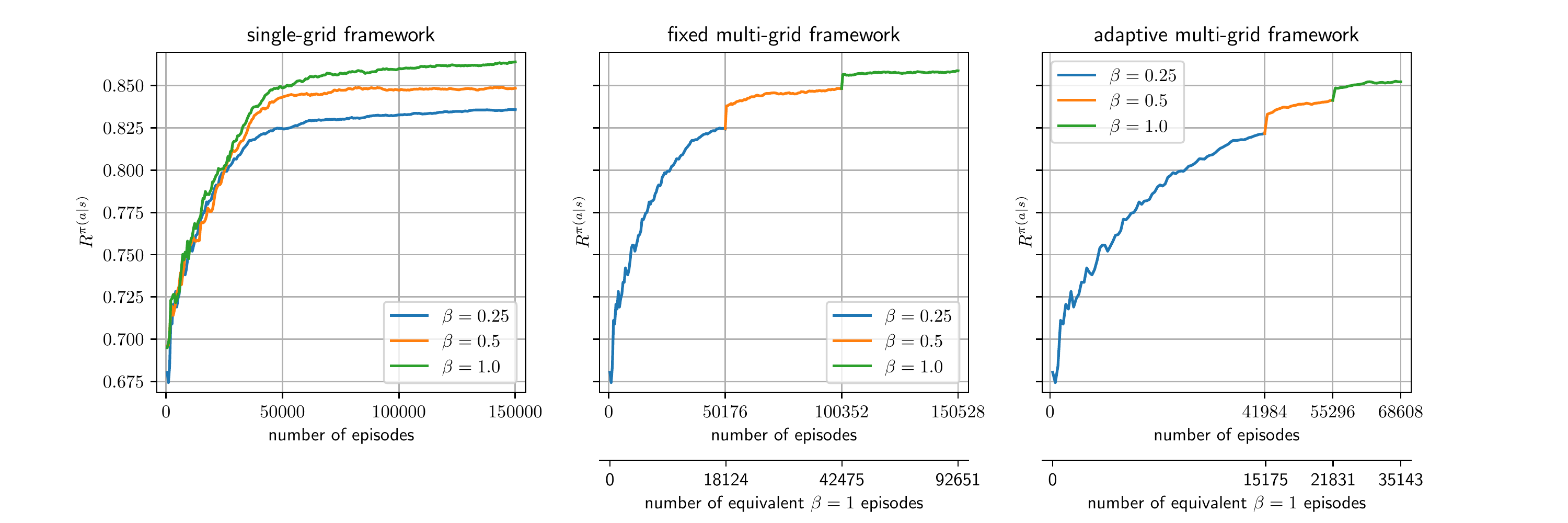}} \\ 
    \end{tabular}
    \caption{learning plots for three distinct seed values for test case 2}
    \label{fig: case_2_seeds}
\end{figure*}

Parameters used for PPO are tabulated in table \ref{tab: ppo_param} which were tuned using trial and error. 
For PPO algorithm, parameters were tuned in order to find least variability in learning plots.
Figures \ref{fig: case_1_seeds} and \ref{fig: case_2_seeds} show learning plots corresponding to three distinct seeds to show the stochasticity of the obtained results.
The DE algorithm's parameters are delineated in table \ref{tab: de_param}. The code repository for both the test cases presented in this paper can be found on the link: \url{https://github.com/atishdixit16/ada\_multigrid\_ppo}.

\begin{table}[ht]
    \caption{PPO algorithm parameters}
    \centering
    \begin{tabular}{l l l l}
        \hline
         & case 1 & case 2\\
        \hline
        number of CPUs, $N$ & 64 & 64 \\
        number of steps, $T$ & 40 & 40 \\
        mini-batch size, $M$ & 16 & 16 \\
        epochs, $K$ & 20 & 20 \\
        discount rate, $\gamma$ & 0.99 & 0.99 \\
        clip range, $\epsilon$ & 0.1 & 0.15 \\
        policy network MLP layers & [93,150,100,80,62] & [35,70,70,50,21] \\
        policy network activation functions  & tanh & tanh\\
        policy network optimizers  & Adam & Adam \\
        learning rate & 3e-6 & 1e-4\\
        \hline
    \end{tabular}
    \label{tab: ppo_param}
\end{table}

\begin{table}[ht]
    \caption{DE algorithm parameters}
    \centering
    \begin{tabular}{l l l l}
        \hline
         & case 1 & case 2\\
        \hline
        number of CPUs  & 64 & 64 \\
        number of iterations & 1024 & 1024 \\
        population size & 310 & 105 \\
        recombination factor & 0.9 & 0.9 \\
        mutation factor & (0.5,1) & (0.5,1) \\
        \hline
    \end{tabular}
    \label{tab: de_param}
\end{table}

%\begin{acknowledgements}
%\end{acknowledgements}

\clearpage

\bibliographystyle{MG}       % Mathematical Geoscience style
{\footnotesize
\bibliography{MG_template}}   % name your BibTeX data base

\begin{thebibliography}{25}
\providecommand{\natexlab}[1]{#1}

\bibitem[Anderlini et~al.(2016)]{anderlini2016control}
Anderlini E, Forehand D~I, Stansell P, Xiao Q, Abusara M (2016) Control of a
  point absorber using reinforcement learning. IEEE Transactions on Sustainable
  Energy 7(4):1681--1690

\bibitem[Anderson and Crawford-Hines(1994)]{anderson1994multigrid}
Anderson C, Crawford-Hines S (1994) Multigrid q-learning. In Technical Report
  CS-94-121, Citeseer

\bibitem[Brouwer et~al.(2001)]{brouwer2001recovery}
Brouwer D, Jansen J, Van~der Starre S, Van~Kruijsdijk C, Berentsen C, et~al.
  (2001) Recovery increase through water flooding with smart well technology.
  In SPE European Formation Damage Conference, Society of Petroleum Engineers

\bibitem[Christie et~al.(2001)]{christie2001tenth}
Christie M~A, Blunt M, et~al. (2001) Tenth {SPE} comparative solution project:
  A comparison of upscaling techniques. In SPE reservoir simulation symposium,
  Society of Petroleum Engineers

\bibitem[Dornheim et~al.(2020)]{dornheim2020model}
Dornheim J, Link N, Gumbsch P (2020) Model-free adaptive optimal control of
  episodic fixed-horizon manufacturing processes using reinforcement learning.
  International Journal of Control, Automation and Systems 18(6):1593--1604

\bibitem[Fachantidis et~al.(2013)]{fachantidis2013transferring}
Fachantidis A, Partalas I, Tsoumakas G, Vlahavas I (2013) Transferring task
  models in reinforcement learning agents. Neurocomputing 107:23--32

\bibitem[Fern{\'a}ndez et~al.(2010)]{fernandez2010probabilistic}
Fern{\'a}ndez F, Garc{\'\i}a J, Veloso M (2010) Probabilistic policy reuse for
  inter-task transfer learning. Robotics and Autonomous Systems 58(7):866--871

\bibitem[Lazaric et~al.(2008)]{lazaric2008transfer}
Lazaric A, Restelli M, Bonarini A (2008) Transfer of samples in batch
  reinforcement learning. In Proceedings of the 25th international conference
  on Machine learning, 544--551

\bibitem[Li and Xia(2015)]{li2015multi}
Li B, Xia L (2015) A multi-grid reinforcement learning method for energy
  conservation and comfort of {HVAC} in buildings. In 2015 IEEE International
  Conference on Automation Science and Engineering (CASE), IEEE, 444--449

\bibitem[Müller and Schüler(2019)]{sebastian_muller_2019_2541735}
Müller S, Schüler L (2019) Geostat-framework/gstools: Bouncy blue

\bibitem[Narvekar et~al.(2016)]{narvekar2016source}
Narvekar S, Sinapov J, Leonetti M, Stone P (2016) Source task creation for
  curriculum learning. In Proceedings of the 2016 international conference on
  autonomous agents \& multiagent systems, 566--574

\bibitem[Pareigis(1996)]{pareigis1996multi}
Pareigis S (1996) Multi-grid methods for reinforcement learning in controlled
  diffusion processes. In NIPS, Citeseer, 1033--1039

\bibitem[Park(2011)]{park2011modeling}
Park K (2011) Modeling uncertainty in metric space. Stanford University

\bibitem[Rabault et~al.(2019)]{rabault2019artificial}
Rabault J, Kuchta M, Jensen A, R{\'e}glade U, Cerardi N (2019) Artificial
  neural networks trained through deep reinforcement learning discover control
  strategies for active flow control. Journal of fluid mechanics 865:281--302

\bibitem[Raffin et~al.(2019)]{stable-baselines3}
Raffin A, Hill A, Ernestus M, Gleave A, Kanervisto A, Dormann N (2019) Stable
  baselines3. \url{https://github.com/DLR-RM/stable-baselines3}

\bibitem[Roseta-Palma and Xepapadeas(2004)]{roseta2004robust}
Roseta-Palma C, Xepapadeas A (2004) Robust control in water management. Journal
  of Risk and Uncertainty 29(1):21--34

\bibitem[Schulman et~al.(2015)]{schulman2015high}
Schulman J, Moritz P, Levine S, Jordan M, Abbeel P (2015) High-dimensional
  continuous control using generalized advantage estimation. arXiv preprint
  arXiv:150602438

\bibitem[Schulman et~al.(2017)]{schulman2017proximal}
Schulman J, Wolski F, Dhariwal P, Radford A, Klimov O (2017) Proximal policy
  optimization algorithms. arXiv preprint arXiv:170706347

\bibitem[Storn and Price(1997)]{storn1997differential}
Storn R, Price K (1997) Differential evolution--a simple and efficient
  heuristic for global optimization over continuous spaces. Journal of global
  optimization 11(4):341--359

\bibitem[Taylor and Stone(2005)]{taylor2005behavior}
Taylor M~E, Stone P (2005) Behavior transfer for value-function-based
  reinforcement learning. In Proceedings of the fourth international joint
  conference on Autonomous agents and multiagent systems, 53--59

\bibitem[Taylor and Stone(2009)]{taylor2009transfer}
Taylor M~E, Stone P (2009) Transfer learning for reinforcement learning
  domains: A survey. Journal of Machine Learning Research 10(7)

\bibitem[van Essen et~al.(2009)]{van2009robust}
van Essen G, Zandvliet M, Van~den Hof P, Bosgra O, Jansen J~D, et~al. (2009)
  Robust waterflooding optimization of multiple geological scenarios. Spe
  Journal 14(01):202--210

\bibitem[Virtanen et~al.(2020)]{2020SciPy-NMeth}
Virtanen P, Gommers R, Oliphant T~E, Haberland M, Reddy T, Cournapeau D,
  Burovski E, Peterson P, Weckesser W, Bright J, {van der Walt} S~J, Brett M,
  Wilson J, Millman K~J, Mayorov N, Nelson A~R~J, Jones E, Kern R, Larson E,
  Carey C~J, Polat {\.I}, Feng Y, Moore E~W, {VanderPlas} J, Laxalde D,
  Perktold J, Cimrman R, Henriksen I, Quintero E~A, Harris C~R, Archibald A~M,
  Ribeiro A~H, Pedregosa F, {van Mulbregt} P, {SciPy 10 Contributors} (2020)
  {{SciPy} 1.0: Fundamental Algorithms for Scientific Computing in Python}.
  Nature Methods 17:261--272

\bibitem[Whitaker(1999)]{whitaker1999single}
Whitaker S (1999) Single-phase flow in homogeneous porous media: Darcy’s law.
  In The method of volume averaging, Springer, 161--180

\bibitem[Ziv and Shimkin(2005)]{ziv2005multigrid}
Ziv O, Shimkin N (2005) Multigrid methods for policy evaluation and
  reinforcement learning. In Proceedings of the 2005 IEEE International
  Symposium on, Mediterrean Conference on Control and Automation Intelligent
  Control, 2005., IEEE, 1391--1396

\end{thebibliography}

%% Non-BibTeX users please use
%\begin{thebibliography}{}
%%
%% and use \bibitem to create references. Consult the Instructions
%% for authors for reference list style.
%\end{thebibliography}

\end{document}